\documentclass[10pt,twocolumn,letterpaper]{article}

\usepackage{cvpr}              

\definecolor{cvprblue}{rgb}{0.21,0.49,0.74}
\usepackage[pagebackref,breaklinks,colorlinks,allcolors=cvprblue]{hyperref}

\def\paperID{*****} 
\def\confName{CVPR}
\def\confYear{2026}

\title{Breaking the Resource Wall: Geometry-Guided Sequence Modeling for Efficient Semantic Segmentation}

\author{
    Sheng-Wei Chan \quad Hsin-Jui Pan \quad Chun-Po Shen\quad Chia-Min Lin \quad Yung-Che Wang \quad Jen-Shiun Chiang\thanks{Corresponding author.} \\
    Department of Electrical and Computer Engineering, Tamkang University\\
    {\tt\small \{412440330, 412440314, 412441064, 170587, 614450012\}@o365.tku.edu.tw} \\
    {\tt\small chiang@mail.tku.edu.tw}
}

\usepackage{graphicx}     
\usepackage{subcaption}  
\usepackage{tabularx}
\usepackage{booktabs} 
\begin{document}
\maketitle

\begin{abstract}
High-performance semantic segmentation has achieved significant progress in recent years, largely driven by increasingly large backbones and higher computational budgets. While effective, such approaches introduce substantial computational overhead and limit accessibility in resource-constrained environments. In this paper, we propose DGM-Net (Directional Geometric Mamba Network), an efficient architecture that enhances modeling capability through structural design without significantly increasing model capacity or computational overhead. We introduce Directional Geometric Mamba (G-Mamba), a linear-complexity $O(N)$ operator that serves as an alternative to conventional context modeling modules such as ASPP (Atrous Spatial Pyramid Pooling) and PPM. To further improve structural awareness in State Space Model (SSM)-based frameworks, we design the DGM-Module, which extracts centripetal flow fields and topological skeleton priors to guide the scanning process and preserve fine-grained boundaries. Without relying on large-scale pretraining or heavy backbone scaling, DGM-Net achieves 80.8\% mIoU within 28k iterations, demonstrating fast convergence and stable scalability. The model further reaches 82.3\% mIoU on the Cityscapes test set and 45.24\% mIoU on ADE20K. Moreover, DGM-Net maintains robust performance under constrained hardware settings (e.g., batch size of 2 on 8GB VRAM), highlighting its efficiency and practicality. These results suggest that incorporating geometric guidance into SSM-based architectures provides an effective and resource-efficient direction for semantic segmentation.
\end{abstract}    
\section{Introduction}

Semantic segmentation is a fundamental task in computer vision, aiming to assign a semantic label to each pixel in an image. With the rapid development of deep learning, the field has undergone several stages of architectural evolution and demonstrated significant impact in applications such as autonomous driving, medical image analysis, and intelligent surveillance.

Early approaches are primarily based on Fully Convolutional Networks~\cite{long2015fully}, which establish dense prediction frameworks through local convolution operations. These methods benefit from strong inductive bias and computational efficiency, allowing them to effectively capture local structures and boundary details. However, due to their limited receptive field, they struggle to model long-range semantic dependencies. To address this limitation, subsequent works introduce multi-scale context modeling techniques, such as atrous convolution and spatial pyramid pooling, as exemplified by DeepLab~\cite{chen2017deeplab, chen2018encoder} and PSPNet~\cite{zhao2017pyramid}. These methods effectively enlarge the receptive field and aggregate multi-scale information, but typically rely on multi-branch structures or dense sampling strategies, leading to increased computational cost and memory consumption.

More recently, Transformer-based models have significantly improved global dependency modeling through self-attention mechanisms~\cite{transformer_vaswani}. With the introduction of Vision Transformers (ViTs)~\cite{vit_dosovitskiy, deit_touvron} and their variants, segmentation performance has reached new levels of accuracy. Representative approaches such as Swin Transformer~\cite{swin_liu}, SegFormer~\cite{xie2021segformer}, and InternImage~\cite{internimage_wang} have become strong baselines in modern semantic segmentation. By directly modeling interactions between arbitrary spatial locations, these methods achieve superior performance in complex scenes. However, the computational complexity of self-attention grows quadratically with input size, resulting in substantial computational and memory overhead for high-resolution images. In practice, these models often rely on large-scale pretraining~\cite{convnext_liu, beit_bao} and significant computational resources to achieve optimal performance, which may limit their applicability in resource-constrained environments.

As an alternative direction, State Space Models (SSMs), such as the recently proposed Mamba~\cite{gu2023mamba}, provide linear computational complexity, enabling efficient long-range dependency modeling. This has attracted increasing attention in recent research, with several visual SSM variants being proposed~\cite{zhu2024visionmamba, liu2024vmamba}. Despite their efficiency, existing SSM-based approaches typically adopt fixed or isotropic scanning strategies, lacking structural awareness of image geometry. This limitation may lead to over-smoothing in fine-grained regions, resulting in degraded performance on boundaries and small objects.

From these observations, it can be seen that semantic segmentation methods still face a fundamental trade-off among local detail modeling, global context understanding, and computational efficiency. CNN-based methods excel at capturing local structures but lack global reasoning capability; Transformer-based methods provide strong global modeling but incur high computational cost; while SSM-based methods offer efficient long-range modeling but remain limited in representing fine-grained structural details.

Motivated by this, we observe that CNNs and SSMs exhibit complementary properties: the former provides precise local structural and boundary information, while the latter enables efficient global context modeling. A natural direction is therefore to combine both paradigms to balance local precision and global understanding. However, a naive integration is insufficient, as it does not address the lack of structural awareness in SSM-based modeling. To this end, we propose DGM-Net, which incorporates explicit geometric priors into the global modeling process to guide feature propagation and context aggregation. Specifically, we introduce a geometry-guided mechanism that enhances the structural awareness of SSMs, improving structural consistency while mitigating semantic leakage. Unlike approaches that rely on large-scale pretraining or increasing model capacity, our method focuses on improving feature propagation efficiency under resource-constrained settings. Experimental results demonstrate that DGM-Net achieves stable training and competitive performance with limited computational resources.

Our contributions can be summarized as follows:

• We propose a geometry-guided SSM framework that introduces explicit structural inductive bias into spatial feature propagation, addressing the inherent limitations of isotropic aggregation in existing SSM-based models.

• We design a geometry-guided scanning mechanism that enables spatially adaptive feature aggregation, effectively reducing semantic leakage and preserving fine-grained structures.

• We demonstrate that our approach achieves a strong performance-efficiency trade-off under resource-constrained settings, without relying on large-scale pretraining or high-memory multi-GPU training, highlighting its practicality for real-world deployment.
\section{Related work}
\label{sec:formatting}

\textbf{Context Modeling and Attention Mechanisms} \\ Semantic segmentation has evolved through a continuous trade-off between capturing global context and maintaining computational efficiency. Early approaches established dense prediction paradigms using fully convolutional networks~\cite{ronneberger2015u,long2015fully}. To overcome the receptive field limitations of FCNs, iconic architectures such as PSPNet~\cite{zhao2017pyramid} and DeepLabV3+~\cite{chen2018encoder} introduced spatial pyramid pooling and atrous convolutions. These seminal works remain the gold standard for efficient context modeling on foundational backbones. Building upon the self-attention mechanism~\cite{vaswani2017attention}, Transformer-based models~\cite{zheng2021rethinking,xie2021segformer, cheng2022masked} and non-local modules~\cite{yue2018compact} set new benchmarks. Within this paradigm, DANet~\cite{fu2019dual} and CCNet~\cite{huang2019ccnet} harvest spatial-channel relationships, while ANN~\cite{zhu2019asymmetric}, EncNet~\cite{zhang2018context}, and Context Prior~\cite{yu2020context} explore semantic priors to regularize modeling. However, a major trend in recent SOTA models is the heavy reliance on massive pre-trained backbones and quadratic complexity $\mathcal{O}(N^2)$~\cite{margatina2019attention}, which creates a significant computational barrier for resource-constrained environments. In contrast, our work seeks to revitalize the efficiency of foundational platforms by proposing a linear-complexity alternative to these heavy attention mechanisms. 

\noindent\textbf{Boundary-Awareness and Geometric Priors} \\ Precise restoration of object geometry, especially for thin structures like poles or fences, is a long-standing challenge. GSCNN~\cite{takikawa2019gated} and PointRend~\cite{kirillov2020pointrend} focus on boundary refinement through decoupled streams or point-based sampling. Regarding geometric representation, Deep Watershed Transform~\cite{bai2017deep} and DeeperLab~\cite{yang2019deeperlab} utilize distance fields, while CentripetalNet~\cite{dong2020centripetalnet} introduces centripetal shifts for object alignment. More recently, Semi-supervised Boundary Segmentation~\cite{chen2025semi} aims to maintain integrity under low annotation costs. While these methods improve boundary quality, they often overlook how geometric priors can actively guide global context aggregation. Our DGM-Module extends these concepts by transforming geometric fields into explicit navigational signals for scanning.

\noindent\textbf{Visual State Space Models}\\ 
Visual State Space Models (SSMs), particularly those based on Mamba~\cite{gu2023mamba}, offer a promising direction for addressing the computational bottlenecks of attention mechanisms. Recent works such as Vim~\cite{zhu2024visionmamba} and VMamba~\cite{liu2024vmamba} demonstrate the feasibility of linear-complexity modeling in vision, with VMamba further introducing cross-scan strategies. RSMamba~\cite{zhao2024rsmamba} extends this paradigm to remote sensing, while GLMamba~\cite{ji2025global} and ECM-Net~\cite{du2026ecmnet} explore hybrid CNN-Mamba architectures. More recent efforts, including Spiral Selective Scan~\cite{shen2026visual} and VCMamba~\cite{munir2025vcmamba}, focus on improving representation coherence.

However, most existing visual SSMs rely on fixed and isotropic scanning patterns that are agnostic to the underlying image structure, which may lead to over-smoothing and semantic leakage. 
In contrast, our G-Mamba introduces a geometry-guided dynamic scanning mechanism that leverages predicted geometric flows to adaptively modulate the scanning process, improving structural consistency while mitigating semantic leakage.

\noindent\textbf{Feature Alignment and Feedback Loops}\\ Spatial misalignment often degrades segmentation masks. SFNet~\cite{li2020semantic} and AlignSeg~\cite{huang2021alignseg} utilize pixel-level offsets to align features. Recently, DFAM~\cite{wen2023dfam}, Dual-path Feature Alignment~\cite{zhang2026dual}, and Adaptive Feature Refinement~\cite{khan2025afrda} further optimized detail preservation. Our GOAD and Feedback Refiner modules are inspired by these techniques but introduce global feedback signals from the Mamba encoder. This closed-loop mechanism allows DGM-Net to precisely repair complex boundaries even in extreme settings (e.g., a single RTX 3060Ti with a batch size of 2), achieving a true balance between performance and accessibility.
\section{Directional Geometric Mamba Network}
\begin{figure*}[t!]
  \centering
  \includegraphics[width=0.9\textwidth]{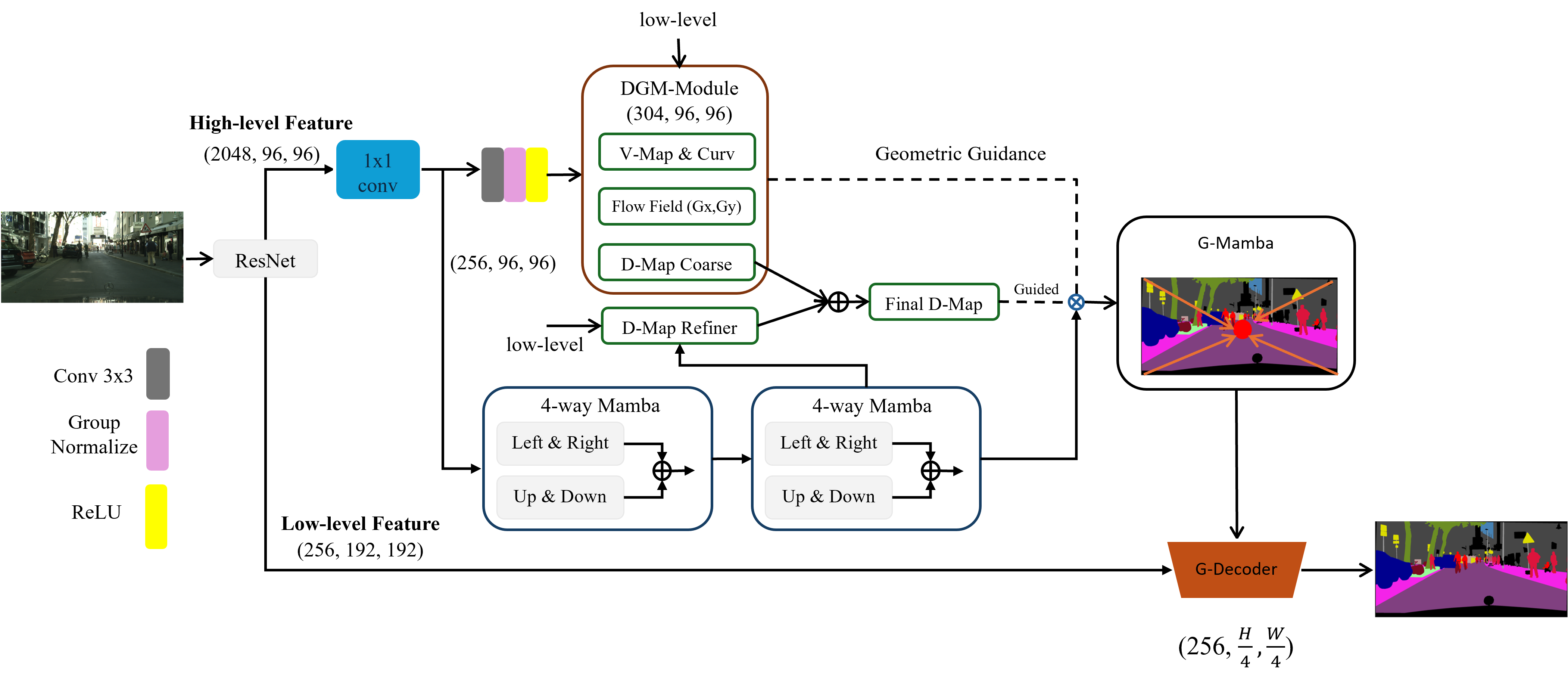}
  \caption{Overview of our proposed DGM-Net architecture. It follows an advanced encoder-decoder paradigm integrated with the Cascade Geometric Navigation SSM and partitions it into four primary stages to ensure structural consistency.}
  \label{fig:main_figure}
\end{figure*}

While State Space Models efficiently capture long-range dependencies, their spatial aggregation is inherently isotropic and lacks explicit structural inductive bias. 
This leads to semantic leakage across object boundaries and over-smoothing of high-frequency structures such as thin objects and edges. To address this limitation, we introduce geometric priors to guide feature propagation. By encoding spatial structure, these priors act as constraints on information flow, restricting feature interaction to geometrically relevant regions. Based on this insight, we propose a geometry-guided scanning mechanism that replaces uniform propagation with spatially adaptive modulation. Instead of treating all locations equally, the model assigns higher importance to structurally critical regions, enabling selective information flow. This can be viewed as importance-weighted sequence modeling, where the effective receptive field is dynamically shaped by geometric cues. As a result, our approach reformulates SSM-based spatial modeling as a geometry-conditioned propagation process, improving structural preservation while maintaining computational efficiency.

\subsection{Overall Architecture}
The proposed DGM-Net follows an advanced encoder-decoder paradigm integrated with the Cascade Geometric Navigation SSM. As illustrated in Fig. \ref{fig:main_figure}, the entire workflow is partitioned into four primary stages to ensure structural consistency:

\begin{itemize}
    \item \textbf{Multi-level Feature Extraction:} We employ a ResNet-101 \cite{he2016deep} backbone with dilated convolutions \cite{yu2015multi} in $Res_{3}$ and $Res_{4}$ stages (output stride 8). This maintains a sufficient receptive field while balancing fine-grained textures and high-level semantics.
    
    \item \textbf{Geometric Prior Generation:} The DGM-Module translates abstract features into explicit topological constraints by predicting three synergistic priors:
    \begin{itemize}
        \item \textit{Centripetal Convergence Field ($V$-map \& Flow):} It defines directional trends pointing toward object centers.
        \item \textit{High-Fidelity Boundary Map ($D$-map):} It captures topological skeletons of slender objects (e.g., poles, fences).
        \item \textit{Curvature-Aware Map ($C$-map):} It highlights regions with significant geometric variations and junctions.
    \end{itemize}
    
    \item \textbf{Cascade G-SSM:} In the three-layer relay Mamba architecture, the first two layers ($L_1, L_2$) establish macro-semantic context, while the third layer ($L_3$) performs asymmetric geometric scanning steered by the predicted Flow Field.
    
    \item \textbf{Decoding and Refinement:} The GOAD module performs pixel-wise warping via grid sampling. A Feedback Refinement mechanism leverages the G-Mamba context to compute a structural residual $\Delta D$, yielding the final high-quality mask.
\end{itemize}

\subsection{DGM-Module}
The DGM-Module translates abstract multi-scale features into explicit topological constraints to guide the subsequent segmentation process. As summarized in Table \ref{tab:geometric_priors}, this module simultaneously predicts four synergistic geometric priors that capture the structural ``DNA'' of the scene. Instead of relying on a single distance metric, we decouple the geometric representation into distinct physical fields to provide both global orientation and local structural anchors.

\begin{table*}[t]
\centering
\caption{The geometric and physical formulations of the priors generated by the DGM-Module.}
\label{tab:geometric_priors}
\renewcommand{\arraystretch}{1.5}
\tabcolsep=12pt 
\begin{tabularx}{\textwidth}{l l X X} 
\toprule
\textbf{Map / Field} & \textbf{Academic Term} & \textbf{Functional Role} & \textbf{Geometric \& Physical Meaning} \\
\midrule
\textbf{V-Map} & Centripetal Potential & Locate structural boundaries and instance centroids. & A normalized scalar potential field indicating distance to the center \\
\textbf{Flow ($\Phi$)} & Directional Flow Field & Guide the spatial convergence direction. & First-derivative gradient vectors pointing towards centroids \\
\textbf{Curv-Map} & Curvature-Aware Map & Highlight turning points and high-curvature regions. & Laplacian-based structural anchors for topological changes \\
\textbf{D-Map} & Morphological Prior & Emphasize hard-to-learn slender structures. & A morphological heat map focusing on boundary skeletons \\
\bottomrule
\end{tabularx}
\end{table*}

\begin{figure}[ht]
  \centering
  \begin{subfigure}[b]{0.48\textwidth}
    \centering
    \includegraphics[width=0.75\textwidth]{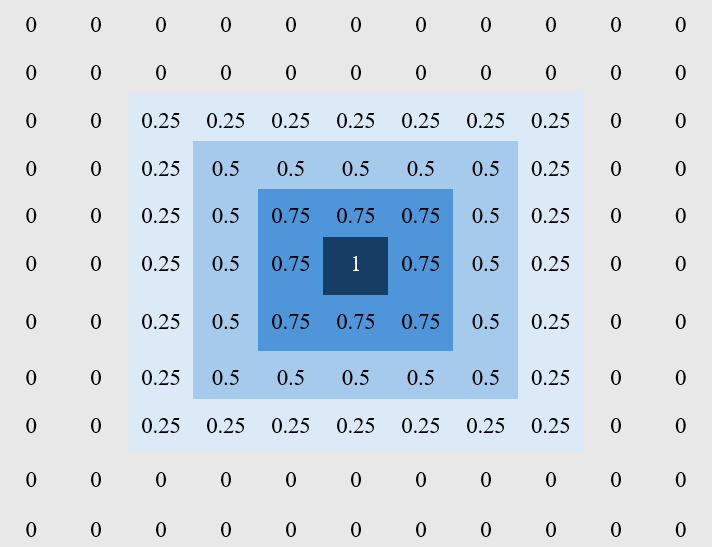}
    \caption{V-Map Visualization}
    \label{fig:vmap_vis}
  \end{subfigure}
  \hfill
  \begin{subfigure}[b]{0.48\textwidth}
    \centering
    \includegraphics[width=0.75\textwidth]{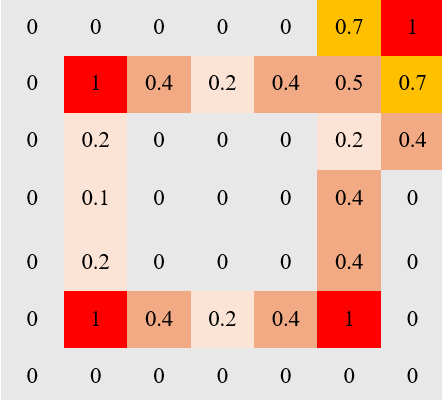}
    \caption{Curv-Map Visualization}
    \label{fig:cmap_vis}
  \end{subfigure}
  \caption{Visualization of geometric priors: (a) $V$-map showing potential fields, and (b) $Curv$-map capturing curvature anchors.}
  \label{fig:priors_visualization}
\end{figure}

\noindent\textbf{Centripetal Flow and Potential} \\We compute the V-Map (see Fig. \ref{fig:vmap_vis}) using a normalized distance transform applied to instance masks, functioning as a scalar potential field to highlight instance centroids. To provide dynamic directional guidance, we extract the Flow Field ($\Phi$) via Sobel filters applied to the V-Map, capturing the first-derivative gradient vectors that point towards object centers. By providing this centripetal guidance, the model can effectively distinguish between adjacent instances and suppress over-segmentation.

\noindent\textbf{Curvature and Structural Anchors}\\ To capture sharp topological transitions such as corners and junctions, we introduce the Curv-Map (see Fig. \ref{fig:cmap_vis}), which is derived using Laplacian derivatives. This acts as a set of static structural anchors that are often lost during standard downsampling. Concurrently, an initial morphological boundary prior, denoted as $D_{coarse}$, is extracted via morphological gradient operations to capture the skeletons of slender objects like poles and fences.

\noindent\textbf{Progressive Feedback Refinement}\\ To prevent boundary fragmentation, we introduce a closed-loop refinement mechanism. In our architecture, $D_{coarse}$ first acts as a soft structural prompt to guide the subsequent Cascade G-Mamba stage. Once the G-Mamba completes its globally-aware spatial scanning, its context-rich features are fed back into a D-Map Refiner to generate a structural residual $\Delta D$. The final high-fidelity Detail Map is thus yielded by:
\begin{equation}
D_{final} = \sigma(D_{coarse} + \Delta D)
\end{equation}
where $\sigma$ represents the Sigmoid activation function.

\begin{figure}[t]
  \centering
  \begin{subfigure}[b]{0.48\textwidth}
    \centering
    \includegraphics[width=0.75\textwidth]{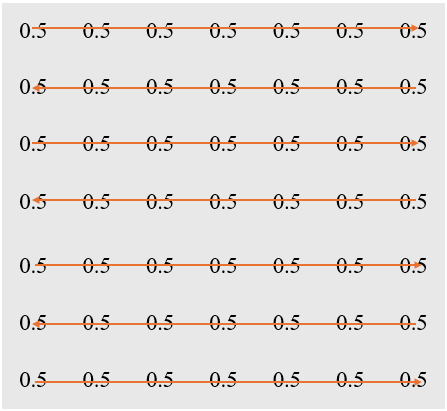}
    \caption{Standard Isotropic Scan}
    \label{fig:mamba_std}
  \end{subfigure}
  \hfill
  \begin{subfigure}[b]{0.48\textwidth}
    \centering
    \includegraphics[width=0.75\textwidth]{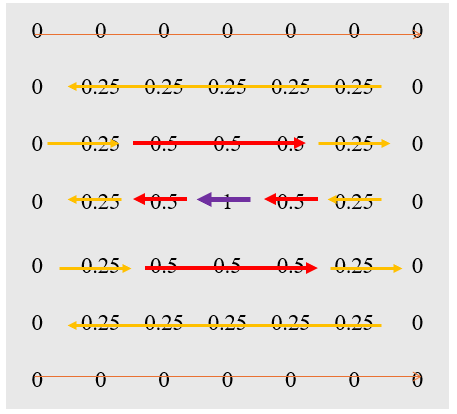}
    \caption{Geometric Navigation Scan}
    \label{fig:mamba_geo}
  \end{subfigure}
  \caption{Comparison of scanning mechanisms. (a) Standard Mamba performs an isotropic scan with uniform spatial priority. (b) Our G-Mamba adaptively modulates the scanning intensity $T$ based on the geometric alignment $\Phi_{dir}$.}
  \label{fig:scanning_comparison}
\end{figure}

\subsection{Cascade Geometric Navigation State-Space Module}
To effectively model global context while maintaining geometric precision, we propose the Cascade G-Mamba (short for Geometry Mamba), a three-layer progressive Mamba architecture. Unlike conventional modules such as ASPP or PPM that rely on multiple parallel dilated convolutions with fixed sampling rates, the G-Mamba reformulates context aggregation as a continuous and geometry-aware sequence modeling process. This architectural shift not only reduces parameter redundancy but also provides a dynamic receptive field that adapts to the underlying object topology, ensuring efficiency on commodity hardware.

\noindent\textbf{Spatial Semantic Grounding and Refinement}\\ The initial two layers ($L_1, L_2$) focus on establishing a robust global semantic distribution. To ensure comprehensive spatial coverage, each block employs a 4-way scanning strategy. Furthermore, to preserve local spatial priors, a depth-wise 2D convolutional prior is applied before the Mamba scan.

\noindent\textbf{Geometric Navigation and Centripetal Alignment}\\ The core innovation lies in the third layer, the DGM-Guided Mamba Block, which functions as a centripetal alignment layer. The fundamental difference between this mechanism and standard scanning is illustrated in Fig. \ref{fig:scanning_comparison}. Unlike the previous layers that perform blind spatial scans with uniform weights (as shown in Fig. \ref{fig:mamba_std}), this block strictly follows the initial structural layout (as shown in Fig. \ref{fig:mamba_geo}).

Specifically, we utilize the centripetal flow field $\Phi$ and the initial coarse map $D_{coarse}$ to compute a soft geometric prompt $T$:
\begin{equation}
T = 1.0 + D_{coarse} \cdot ReLU(\Phi_{dir})
\end{equation}
Physically, this geometric modulation allows the SSM to concentrate its recurrent hidden states on structurally critical regions. By filtering irrelevant spatial noise through the prompt $T$, DGM-Net achieves a more sparse and focused context aggregation compared to traditional isotropic (blind) scans, significantly accelerating convergence. Prior to the SSM scan, the input features are spatially reweighted by the geometric prompt $T$. The modulated feature $F'_{in}$ can be formulated as:
\begin{equation}
F'_{in} = F_{in} \odot T
\end{equation}
where $\odot$ denotes the Hadamard product (element-wise multiplication). This mechanism effectively suppresses ``semantic leakage''—the undesirable diffusion of features across object boundaries—achieving a superior balance between global reasoning and topological consistency.

\subsection{Geometric Offset Alignment Decoder (GOAD)}

To mitigate the spatial misalignment between high-level semantic features and low-level structural details, we introduce Geometric Offset Alignment Decoder (GOAD). While traditional decoders often suffer from blurred boundaries due to the resolution gap between stages, our GOAD module leverages the refined geometric priors $\Phi$ and $D_{final}$ to perform pixel-wise feature warping.

\paragraph{Refined Distance Map}
The final geometric boundary map $D_{final}$ is obtained via a residual refinement process:
\begin{equation}
D_{final} = \sigma(D_{coarse} + \Delta D)
\end{equation}
where $\Delta D$ is predicted by a lightweight refinement head conditioned on fused semantic and low-level features, and $\sigma(\cdot)$ denotes the sigmoid function. This formulation allows the model to correct coarse boundary predictions using global contextual reasoning. 

\paragraph{Geometric Offset Field}
We define a geometry-aware offset field $\Delta$ as:
\begin{equation}
\Delta = \Phi \odot D_{final} \odot \alpha
\end{equation}
where:
\begin{itemize}
    \item where $\odot$ denotes the Hadamard product (element-wise multiplication).
    \item $\Phi \in \mathbb{R}^{2 \times H \times W}$ is the predicted flow field (normalized directional vectors),
    \item $D_{final} \in [0,1]^{H \times W}$ acts as a spatial confidence mask,
    \item $\alpha \in [0, \alpha_{max}]^{H \times W}$ is a learnable dynamic scaling factor.
\end{itemize}

In practice, $\alpha$ is implemented as:
\begin{equation}
\alpha = 0.2 \cdot \sigma(\psi(\Phi))
\end{equation}
where $\psi(\cdot)$ is a shallow convolutional module. This formulation is conceptually related to deformable feature alignment \cite{dai2017deformable}.

\paragraph{Aligned Sampling Grid}
Given a normalized base grid $\mathcal{G}_{base} \in [-1,1]^{H \times W \times 2}$, the aligned grid is constructed as:
\begin{equation}
\mathcal{G}_{aligned} = \mathrm{clip}(\mathcal{G}_{base} + \Delta,\ -1,\ 1)
\end{equation}

\paragraph{Feature Warping}
The aligned low-level feature $F_{aligned}$ is obtained using bilinear sampling:
\begin{equation}
F_{aligned}(p) = \sum_{q \in \mathcal{N}(\mathcal{G}_{aligned}(p))} 
w(p, q) \cdot F_{low}(q)
\end{equation}
where $\mathcal{N}(\cdot)$ denotes the 4-neighbor interpolation set and $w(p,q)$ are bilinear interpolation weights. This operation corresponds to the \texttt{grid\_sample} function in PyTorch \cite{jaderberg2015spatial}, enabling differentiable spatial transformation.

\paragraph{Spatial Gating Mechanism}
To enforce boundary adherence, we apply a multiplicative spatial gating:
\begin{equation}
F_{up} = F_{up} + (F_{up} \odot D_{final})
\end{equation}

\begin{equation}
(F \odot D)(i,j,c) =
\begin{cases}
F(i,j,c) \cdot D(i,j), & \text{if valid pixel} \\
0, & \text{otherwise}
\end{cases}
\end{equation}

This formulation emphasizes features located on predicted structural boundaries while suppressing irrelevant regions.

By employing $D_{final}$ as a spatial gate, the GOAD module ensures that low-level structural details are strictly confined within the predicted geometric boundaries. This mechanism effectively ``paints'' sharp object silhouettes and prevents semantic over-smoothing. Compared to naive upsampling, this design explicitly enforces geometry-aware alignment and reduces boundary ambiguity.

\subsection{Loss Function}

To optimize DGM-Net, we define a composite multi-task objective:
\begin{equation}
\mathcal{L}_{total} = \mathcal{L}_{seg} + \gamma_{geo}(\mathcal{L}_{vg} + \mathcal{L}_{d}) + 0.4\mathcal{L}_{aux}
\end{equation}

\paragraph{Segmentation Loss}
The segmentation objective consists of:
\begin{equation}
\mathcal{L}_{seg} = \mathcal{L}_{CE}^{OHEM} + 0.8 \mathcal{L}_{Lovasz} + 0.1 \mathcal{L}_{boundary}
\end{equation}
where:
\begin{itemize}
    \item $\mathcal{L}_{CE}^{OHEM}$ \cite{shrivastava2016training} selects the top-$k$ hardest pixels based on confidence, 
    \item $\mathcal{L}_{Lovasz}$ \cite{berman2018lovasz} directly optimizes IoU,
    \item $\mathcal{L}_{boundary}$ reweights pixels using the ground-truth $D$ map to emphasize boundaries.
\end{itemize}

\paragraph{Geometric Loss}
The geometric supervision consists of:
\begin{equation}
\mathcal{L}_{vg} = \mathcal{L}_{MSE}(V,\hat{V}) + \mathcal{L}_{MSE}(\Phi,\hat{\Phi}) + \mathcal{L}_{MSE}(C,\hat{C}) + 0.5 \mathcal{L}_{TV}(\Phi)
\end{equation}
where $\mathcal{L}_{TV}$ enforces spatial smoothness on the flow field.

\paragraph{Distance Map Loss}
\begin{equation}
\mathcal{L}_{d} = \mathrm{BCE}_{weighted}(D_{pred}, D_{gt})
\end{equation}
with a positive class weight of $20.0$ to emphasize thin structures such as poles and fences.

\paragraph{Auxiliary Loss}
\begin{equation}
\mathcal{L}_{aux} = \mathcal{L}_{CE}(P_{aux}, Y)
\end{equation}
which stabilizes intermediate supervision.

The coefficient of $0.4$ is empirically determined to balance auxiliary guidance without overwhelming the main objective.

\paragraph{Dynamic Geometry Weight}
The geometric weight $\gamma_{geo}$ follows a linear decay schedule:
\begin{equation}
\gamma_{geo} = 2.0 \cdot (1 - t) + 0.2
\end{equation}
where $t \in [0,1]$ denotes normalized training progress. This design encourages early geometric grounding and later semantic refinement.
\section{Experiments}
In this section, we provide a comprehensive quantitative and qualitative evaluation of DGM-Net on the Cityscapes dataset. Unless otherwise specified, all reported results are based on a 28k-iteration training schedule, which serves as the default setting throughout this section. This allows us to consistently evaluate model performance under a fast convergence regime. We assess the effectiveness of the proposed method through comparisons with representative approaches and detailed ablation studies.

\subsection{Implementation Details}
Our model is optimized using the AdamW optimizer \cite{loshchilov2017decoupled} with an initial learning rate of $1 \times 10^{-4}$ and a poly decay power of 0.9. To ensure stable optimization under single-GPU memory constraints, we adopt a mini-batch size of 4. This strategy reduces gradient variance and improves convergence stability while maintaining low memory consumption. It is worth noting that operations in the forward pass (e.g., Batch Normalization) are still computed based on the mini-batch size of 4. The model is trained for 150 epochs with a crop size of $768 \times 768$ and output stride of 8 on a single NVIDIA RTX 5080 GPU. 

To provide a robust foundation for feature extraction, the backbone (ResNet-101) is initialized with weights pre-trained on the ImageNet-1K dataset \cite{russakovsky2015imagenet}. To emphasize the accessibility and efficiency of our architecture, we do not utilize any further pre-training on large-scale segmentation datasets such as MS-COCO \cite{lin2014microsoft} or Mapillary Vistas \cite{neuhold2017mapillary}. We rely primarily on the official Cityscapes dataset \cite{cordts2016cityscapes} to evaluate the intrinsic performance of the proposed Mamba-based modules.

During training, we apply a robust augmentation suite, including random scaling with a ratio between 0.5 and 2.0, random horizontal flipping, and random color jittering. For performance evaluation, we report results on the Cityscapes validation set. In our best-performing configurations, we utilize Multi-Scale Testing with scaling factors of $\{0.5, 0.75, 1.0, 1.25, 1.5\}$ and horizontal flipping. While the primary benchmarks are established on an RTX 5080, we further evaluate the model’s resilience under stricter 8GB VRAM constraints (e.g., RTX 3060Ti) in Sec. 4.5 to demonstrate its accessibility.

\begin{table*}[t]
\centering
\caption{Quantitative results on Cityscapes validation and test sets.
(B): models trained on both the train and validation sets.
(C): models trained with additional coarse annotations (coarse+fine).}
\label{tab:sota_hierarchical}
\small
\begin{tabular}{lcccccccc}
\toprule
 & & & & \multicolumn{2}{c}{Val mIoU (\%)} & \multicolumn{3}{c}{Test mIoU (\%)} \\
\cmidrule(r){5-6} \cmidrule(l){7-9}
Model & Backbone & Params & GFLOPs & SS & MST+FF & Base & (B) & (C) \\ 
\midrule
DeepLabV3\cite{chen2017rethinking}  & ResNet-101 & 60.22 & 254.0 & 77.23 & 79.3 & - & - & 81.3\\
DeepLabV3+ & ResNet-101 & 84.74 & 348.0 & - & - & 79.55 & - & 82.10 \\
DANet      & ResNet-101 & 66.47 & 289.0 & 78.8 & 81.5 & - & 81.50 & - \\
GSCNN\cite{takikawa2019gated}      & WideResNet-38 & 129.17 & - & 80.8 & - & - & 82.8 & - \\
CCNet      & ResNet-101 & 66.13 & 276.0 & 80.5 & 81.4 & - & 81.90 & - \\
PSPNet     & ResNet-101 & 65.60 & 256.0 & - & - & 78.40 & - & 80.20 \\
PSANet\cite{zhao2018psanet}     & ResNet-101 & 78.13 & 264 & 78.6 & 79.77 & - & 80.10 & 81.40 \\
ANNNet     & ResNet-101 & 62.86 & 348.0 & 80.5 & 81.3 & - & 81.3 & - \\
DNL     & ResNet-101 & 71.49 & 343.0 & 80.5 & - & - & 82 & - \\
OCRNet\cite{yuan2020object}     & ResNet-101 & 70.41 & 325.0 & 81.8 & - & - & 82.4 & - \\
\midrule
\textbf{DGM-Net} & \textbf{ResNet-101} & \textbf{53.00} & \textbf{224.1} & \textbf{80.8} & \textbf{82.3} & \textbf{80.50} & \textbf{81.64} & \textbf{82.3} \\ 
\bottomrule
\end{tabular}
\end{table*}

\subsection{Performance Analysis}
In this section, we provide a comprehensive analysis of the training dynamics, scaling potential, and real-time efficiency of DGM-Net.

\noindent\textbf{Convergence Efficiency}\\ DGM-Net achieves 80.8\% mIoU on the Cityscapes validation set within 28k iterations, showing stable convergence behavior. This ``fast-track'' training capability is particularly advantageous for research environments with limited GPU resources, significantly reducing the computational overhead and time cost of algorithmic iteration. When applying this 28k-iteration schedule to the combined train+val set, DGM-Net achieves 81.64\% mIoU on the official test set, demonstrating that our geometric priors effectively leverage additional data without requiring extended training cycles.\\
\noindent\textbf{Scaling Potential}\\ To demonstrate that our hybrid architecture is not capacity-capped by its efficient design, we extended the training to a full 90k-iteration schedule. Under this configuration, the model steadily scales to a peak performance of 81.6\% mIoU (Single-Scale). This validates that the Directional Geometric Mamba maintains high learning capacity even after the initial rapid convergence phase.\\
\noindent\textbf{Inference Speed and Efficiency}\\
Efficiency is a core strength of the DGM-Net architecture. When evaluated on a single NVIDIA RTX 5080 GPU, our model processes $1024 \times 1024$ resolution inputs at 53.48 FPS. This high throughput, combined with a computational footprint of 896 GFLOPs at this resolution, makes DGM-Net highly suitable for real-time deployment in high-stakes domains such as autonomous driving.\\
\noindent\textbf{Multi-Scale Testing Gains}\\ We further evaluate the robustness of the learned representations using the MST strategy described in Sec. 4.1. By incorporating multi-scale inference and horizontal flipping, the 28k-iteration model is boosted to 82.3\% mIoU, representing a significant gain of +1.5\% over the single-scale baseline. This improvement confirms that the geometric priors extracted by the DGM-Module provide strong structural guidance that generalizes effectively across varying input resolutions.

\subsection{Comparison with Representative Methods}
As shown in Table~\ref{tab:sota_hierarchical}, DGM-Net achieves 80.8\% mIoU on Cityscapes validation with 53M parameters and 224.1 GFLOPs. Although the absolute accuracy does not surpass the strongest prior methods, the result demonstrates a favorable efficiency--accuracy trade-off under a short 28k-iteration training schedule.

\noindent\textbf{Comparison with ASPP-based Designs}\\
To further validate the effectiveness of the proposed DGM-based architecture, we compare it with the widely-used Atrous Spatial Pyramid Pooling (ASPP) module, which serves as a standard component for multi-scale context modeling in many segmentation frameworks.

As shown in Table~\ref{tab:compare_aspp}, replacing ASPP with our DGM + G-Mamba design leads to a more favorable trade-off between accuracy and computational efficiency. Specifically, while ASPP-based models rely on multi-branch dilated convolutions to enlarge the receptive field, our method achieves comparable or superior performance with significantly reduced computational overhead and parameter count.

These results suggest that the proposed DGM-module, when integrated with G-Mamba, can effectively capture both local geometric structures and long-range dependencies without resorting to heavy multi-scale pooling operations. This suggests that our approach provides an effective and lightweight alternative to conventional ASPP-based designs.

\begin{table}[htbp]
\centering
\small
\setlength{\tabcolsep}{5pt}
\caption{Comparison with representative methods on the Cityscapes validation set. Training requirements are summarized based on commonly reported configurations. Training requirements are summarized based on commonly reported settings in the literature.}
\label{tab:cityscapes_comparison}

\resizebox{\linewidth}{!}{
\begin{tabular}{@{}llcc@{}}
\toprule
Method & Backbone & mIoU (\%) & Training Requirement \\ \midrule
SegFormer \cite{xie2021segformer} & MiT-B2 & 81.1 & Large-scale pretraining \\
SegFormer \cite{xie2021segformer} & MiT-B5 & 84.2 & Large-scale pretraining \\
Swin-S \cite{liu2021swin} & Swin-Small & 82.1 & Multi-GPU / high-memory training \\
Mask2Former \cite{cheng2022masked} & Swin-L & 84.3 & Multi-GPU / large-scale pretraining \\
Vim \cite{zhu2024visionmamba} & Vim-L & 80.6 & High-memory training \\
\textbf{DGM-Net (Ours)} & \textbf{ResNet-101} & \textbf{81.6} & \textbf{Single GPU (16GB)} \\
\bottomrule
\end{tabular}
}

\end{table}

We further compare DGM-Net with recent state-of-the-art transformer-based approaches to contextualize its performance (see Table~\ref{tab:cityscapes_comparison}). While transformer-based models achieve strong absolute performance, they typically rely on large-scale pretraining (e.g., ImageNet-22K) and are commonly trained under high-memory, multi-GPU settings, which may limit their accessibility in resource-constrained environments.

In contrast, DGM-Net is explicitly designed to operate under limited computational budgets. Despite using a single GPU with 16GB VRAM and without large-scale pretraining, our method achieves competitive performance. This comparison highlights a practical trade-off between accuracy and computational accessibility, demonstrating that high-quality semantic segmentation can be achieved without excessive computational resources, while maintaining competitive performance.

\begin{figure}[t!]
\centering
\includegraphics[width=\linewidth]{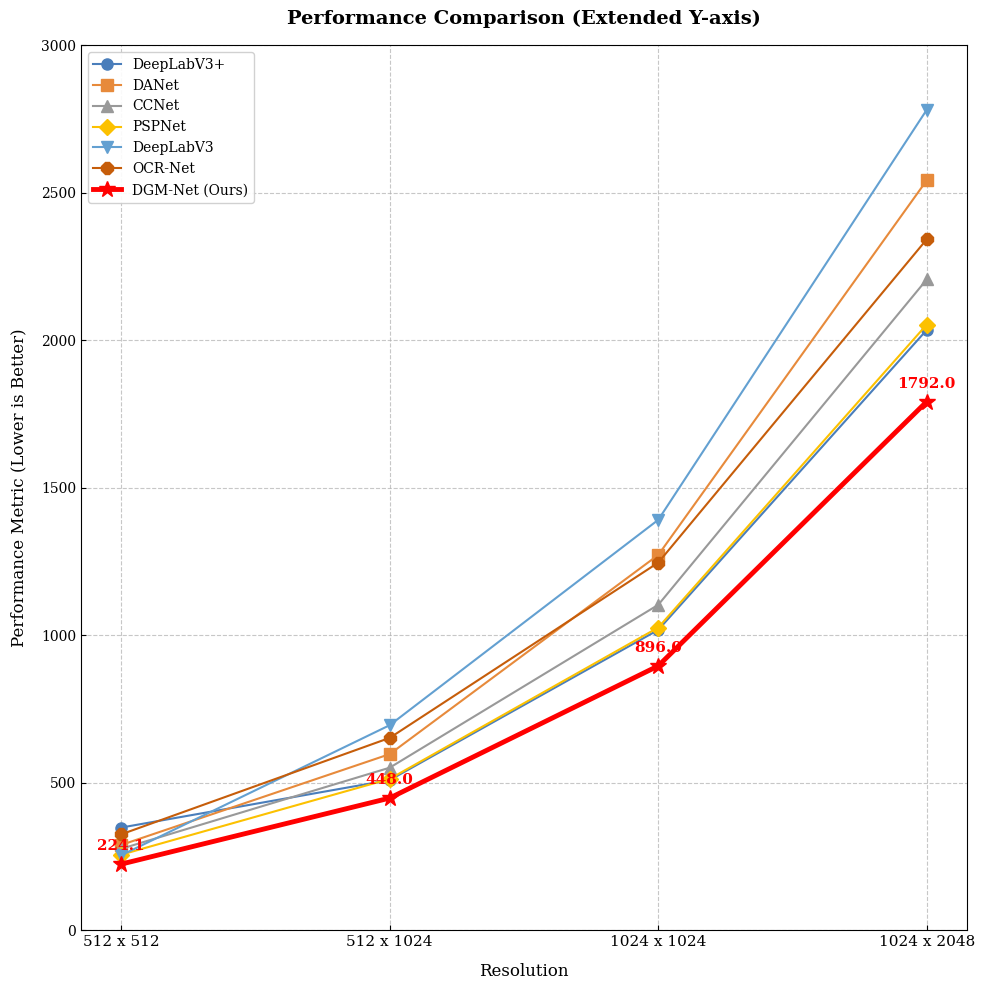}
\caption{Computational complexity (GFLOPs) scaling across varying input resolutions. While classic attention-based models and the efficient OCR-Net~\cite{yuan2020object} exhibit steep and super-linear growth, DGM-Net maintains a highly favorable and linear scaling profile, making it uniquely suited for high-resolution inference.}
\label{fig:complexity_plot}
\end{figure}

\noindent\textbf{Computational Complexity and Scaling} \\
We analyze the architectural efficiency in Table~\ref{tab:gflops_res} and visualize the scaling trends in Figure~\ref{fig:complexity_plot}. DGM-Net maintains a superior accuracy-to-latency trade-off, outperforming traditional CNN-based attention mechanisms. 

Notably, even when compared against OCR-Net~\cite{yuan2020object}—which is recognized for its optimized context aggregation—DGM-Net demonstrates a much flatter growth curve as resolution increases. At the standard $1024 \times 2048$ resolution, OCR-Net's computational demand escalates to 2,342 GFLOPs, whereas DGM-Net remains at 1,792 GFLOPs, representing a 23.5\% reduction in total operations. This linear scaling behavior confirms that the geometry-guided Mamba mechanism effectively bypasses the quadratic or heavy-branch overhead found in current state-of-the-art architectures, ensuring high-throughput performance even at ultra-high resolutions.

\begin{table*}[t]
\centering
\small
\caption{Computational complexity (GFLOPs) vs. input resolutions ($H \times W$). All models are measured using a single input batch. All models are measured using a single input batch. GFLOPs vary with input resolution (e.g., 224.1 GFLOPs at 512 $\times$ 512 and 896 GFLOPs at 1024 $\times$ 1024 for DGM-Net).}
\label{tab:gflops_res}
\begin{tabular}{l cccc}
\toprule
Model & $512\times512$ & $512\times1024$ & $1024\times1024$ & $1024\times2048$ \\ 
\midrule
DeepLabV3+ & 348.0 & 509.0 & 1018.0 & 2035.0 \\
DANet      & 289.0 & 597.0 & 1272.0 & 2542.0 \\
CCNet      & 276.0 & 551.0 & 1103.0 & 2206.0 \\
PSPNet     & 256.0 & 513.0 & 1026.0 & 2053.0 \\
DeepLabV3  & 254.0 & 695.0 & 1391.0 & 2781.0 \\
OCRNet     & 325.0 & 652.0 & 1246.0 & 2342.0 \\
\midrule
\textbf{DGM-Net (Ours)} & \textbf{224.1} & \textbf{448.0} & \textbf{896.0} & \textbf{1792.0} \\ 
\bottomrule
\end{tabular}
\end{table*}

\noindent\textbf{Computational Complexity}\\ We analyze the efficiency in Table~\ref{tab:gflops_res}. DGM-Net maintains a superior accuracy-to-latency trade-off, outperforming traditional CNN-based attention mechanisms.

\subsection{Ablation Study}
\textbf{Mamba Configuration Search} We conduct a search for the optimal block configuration in the GN-SSM. As shown in Table~\ref{tab:mamba_search}, the Serial (2*Mamba + 1*G-Mamba) configuration achieves the peak mIoU.

\begin{table}[h]
\centering
\small
\caption{Ablation on Mamba/G-Mamba configurations. (All with DGM-Module).}
\label{tab:mamba_search}
\begin{tabular}{llc}
\toprule
Configuration & Structure & mIoU (\%) \\ 
\midrule
1 * Mamba & Baseline & 79.60 \\
1 * G-Mamba & Baseline & 80.11 \\
2 * G-Mamba & Baseline & 80.31 \\
3 * G-Mamba & Baseline & 80.57 \\ 
\midrule
1 * Mamba + 1 * G-Mamba & Serial & 80.47 \\
\textbf{2 * Mamba + 1 * G-Mamba} & \textbf{Serial} & \textbf{80.80} \\
3 * Mamba + 1 * G-Mamba & Serial & 80.72 \\ 
\midrule
1 * Mamba + 1 * G-Mamba & Parallel & 80.04 \\
2 * Mamba + 1 * G-Mamba & Parallel & 80.24 \\
3 * Mamba + 1 * G-Mamba & Parallel & 80.33 \\
\bottomrule
\end{tabular}
\end{table}

\noindent\textbf{Effect of Incremental Components}\\ To quantify the contribution of each proposed module, we perform an incremental ablation study starting from a vanilla ResNet-101 baseline. As summarized in Table~\ref{tab:incremental_gain}, each component provides a steady improvement in mIoU.

\noindent\textbf{Robustness under Extreme Memory Constraints}\\ 
Table~\ref{tab:vram_battle} presents a comparative analysis under a memory-constrained setting using a single 8GB VRAM GPU. Under such constraints, conventional architectures such as DeepLabV3+ and PSPNet require reducing the input crop size (e.g., $641 \times 641$) to avoid out-of-memory issues, which may limit their effective receptive field and contextual coverage. In contrast, the linear memory complexity of the proposed G-Mamba allows DGM-Net to maintain a larger crop size ($768 \times 768$) under the same hardware budget. This enables richer spatial context modeling while preserving stable performance. We note that all models are evaluated under memory-feasible configurations to ensure a fair comparison.

\begin{itemize}
    \item \textbf{Larger Effective Receptive Field:} 
    By supporting a larger crop size under the same memory constraint, DGM-Net benefits from richer spatial context, which is particularly important for urban scene understanding.

    \item \textbf{Stable Single-Scale Performance:} 
    DGM-Net achieves competitive performance under single-scale inference without relying on multi-scale testing, indicating that the geometry-guided scanning mechanism provides consistent structural modeling.

    \item \textbf{Robustness to Memory Constraints:} 
    While performance of all methods is affected under strict memory settings, DGM-Net maintains relatively stable accuracy, demonstrating its robustness in resource-constrained environments.
\end{itemize}

\begin{table}[h]
\centering
\caption{Comparison under strict hardware constraints (Single 8GB VRAM GPU). DGM-Net effectively breaks the ``resource wall'' by enabling larger crop sizes without memory overflow.}
\label{tab:vram_battle}
\small 
\tabcolsep=4pt 
\begin{tabular}{lcccc}
\toprule
Model (Backbone: R-101) & Max Crop & BS & MST & Val mIoU \\ 
\midrule
DeepLabV3+ \cite{chen2018encoder} & 641 $\times$ 641 & 2 & \checkmark & 77.43 \\
DeepLabV3+ \cite{chen2018encoder} & 641 $\times$ 641 & 2 & $\times$   & 69.42 \\
\midrule
PSPNet \cite{zhao2017pyramid}    & 641 $\times$ 641 & 2 & \checkmark & 77.31 \\
PSPNet \cite{zhao2017pyramid}    & 641 $\times$ 641 & 2 & $\times$   & 67.3 \\
\midrule
\textbf{DGM-Net (Ours)} & \textbf{768 $\times$ 768} & \textbf{2} & $\times$ & \textbf{77.45} \\ 
\textbf{DGM-Net (Ours)} & \textbf{768 $\times$ 768} & \textbf{2} & \checkmark & \textbf{79.8} \\
\bottomrule
\end{tabular}
\end{table}

\noindent\textbf{Effect of Feedback Refinement}\\ To verify the necessity of the feedback mechanism in our DGM-Module, we conduct a controlled experiment on the D-Map generation process. As shown in Table~\ref{tab:incremental_gain}, excluding the Feedback Refiner (using only $D_{coarse}$) results in a performance drop to 79.87\% mIoU. By incorporating the structural residual $\Delta D$ from global Mamba context, the accuracy improves to 80.80\% (+0.93\%). This confirms that the closed-loop refinement is crucial for repairing complex topological details.

\noindent\textbf{Synergy of Hybrid Scanning} \\
Our hierarchical design is critical for achieving rapid convergence. The initial layers ($L_1, L_2$) utilize symmetric 4-way scanning to establish a generic global semantic field, essentially answering ``what'' objects are present. The final G-Mamba layer ($L_3$) then acts as a precision filter, leveraging the DGM-guided flow to answer ``where'' the boundaries reside. This division of labor allows DGM-Net to bypass the exhaustive spatial groping phase typical of isotropic attention mechanisms, achieving a stable 80.8\% mIoU in less than one-third of the training cycles required by conventional models.

\begin{table}[h]
\centering
\small
\caption{Step-by-step incremental performance gain of DGM-Net components on Cityscapes.}
\label{tab:incremental_gain}
\begin{tabular}{lcc}
\toprule
Component / Modification & mIoU (\%) & Gain ($\Delta$) \\ 
\midrule
Baseline (ResNet-101) & 74.20 & - \\
+ DGM-Module & 77.50 & +3.30 \\
+ Standard Mamba & 79.45 & +1.95 \\
+ G-Mamba (Single) & 80.11 & 0.66 \\
+ Cascade G-Mamba (2+1 Serial) & 80.80 & +0.69 \\ 
\midrule
\textbf{+ Multi-Scale Testing (MST+FF)} & \textbf{82.30} & \textbf{+1.50} \\ 
\bottomrule
\end{tabular}
\end{table}

\begin{table}[t]
\centering
\caption{Comparison with ASPP-based and Mamba-based designs.}
\label{tab:compare_aspp}
\begin{tabular}{lc}
\hline
Method & mIoU (\%) \\
\hline
DGM + ASPP & 78.6 \\
DGM + Mamba & 79.5 \\
DGM + G-Mamba & 80.11 \\
DGM + Cascade G-Mamba & 80.8 \\
\hline
\end{tabular}
\end{table}

\subsection{Training Efficiency and Hardware}
\noindent\textbf{Convergence Efficiency and Scaling Potential}\\ 
As summarized in Table~\ref{tab:convergence}, DGM-Net exhibits a rapid convergence rate. Designed specifically for resource-constrained environments, the model achieves a highly competitive 80.8\% mIoU at an accelerated 28k-iteration schedule. This fast-track training capability is particularly advantageous for research laboratories with limited GPU hours, drastically reducing the carbon footprint and time cost of algorithmic iteration. Furthermore, to demonstrate that our hybrid architecture is not capacity-capped by its lightweight design, we extended the training to a standard 90k iteration schedule. Under this full schedule, the model steadily scaled to a peak performance of 81.6\% mIoU. This validates that the Directional Geometric Mamba maintains high learning capacity even after the initial rapid convergence phase, providing both a fast baseline for efficient research and the potential to compete at higher-end performance tiers.

\begin{table}[h]
\centering
\small
\caption{Performance vs. Training Iterations.}
\label{tab:convergence}
\begin{tabular}{lcc}
\toprule
Schedule & Iterations & mIoU (\%) \\ 
\midrule
Fast (Ours) & 28,000 & 80.8 \\
Full (Ours) & 90,000 & \textbf{81.6} \\ 
\bottomrule
\end{tabular}
\end{table}

\noindent\textbf{Effectiveness on Slender and Boundary-Sensitive Structures} \\
To further understand the underlying reasons for the performance gains of DGM-Net, we analyze the per-class mIoU improvements during the training progression from the fast 28k schedule to the full 90k schedule. As demonstrated in Table \ref{tab:per_class_slender}, the most significant accuracy boosts are not observed in massive background classes (e.g., road, sky), but rather in geometrically complex and thin structures.

Specifically, we observe substantial absolute gains in categories that are notoriously difficult to segment: wall (+2.50\%), fence (+1.87\%), pole (+1.20\%), and traffic light (+1.09\%). This per-class breakdown strongly supports our architectural motivation. By introducing the morphological boundary prior (D-Map) and the centripetal scanning mechanism (G-Mamba), DGM-Net effectively guides the network to focus on topological skeletons. This successfully mitigates the fracturing and semantic leakage issues commonly faced by traditional isotropic attention methods when dealing with slender objects.

\begin{table}[htbp]
\centering
\caption{Per-class mIoU improvements on geometrically challenging structures. The model exhibits significant gains on slender and boundary-sensitive classes when extending the training schedule, validating the effectiveness of our geometric priors.}
\label{tab:per_class_slender}
\resizebox{\columnwidth}{!}{%
\begin{tabular}{lccccc}
\toprule
\textbf{Configuration} & \textbf{Wall} & \textbf{Fence} & \textbf{Pole} & \textbf{Traffic Light} & \textbf{Overall mIoU} \\
\midrule
DGM-Net (28k Iters) & 58.60 & 61.83 & 69.30 & 74.01 & 80.80 \\
DGM-Net (90k Iters) & \textbf{61.10} & \textbf{63.70} & \textbf{70.50} & \textbf{75.10} & \textbf{81.60} \\
\midrule
\textit{Absolute Gain} & \textit{+2.50} & \textit{+1.87} & \textit{+1.20} & \textit{+1.09} & \textit{+0.80} \\
\bottomrule
\end{tabular}%
}
\end{table}

\noindent\textbf{Hardware Constraints}\\
Table~\ref{tab:hardware_specs} evaluates the performance of the proposed method under different hardware settings. All experiments are conducted using single-GPU setups with limited memory (16GB VRAM), without relying on high-memory or server-grade devices.

Despite these constraints, the proposed method achieves strong performance, reaching 80.8\% mIoU on an RTX 5080. Furthermore, when evaluated under more constrained memory settings (e.g., 8GB GPUs), the model still maintains competitive accuracy, with only moderate performance degradation (77.97\% and 77.45\%, respectively).

These results indicate that the effectiveness of the proposed architecture does not depend on large-scale computational resources. Instead, it enables stable and high-quality segmentation performance even in resource-limited environments. This makes the method particularly suitable for laboratories or research settings with constrained hardware budgets, while still achieving performance comparable to, or even approaching, previously established strong baselines.

\begin{table}[h]
\centering
\small
\caption{Hardware environment and training constraints.}
\label{tab:hardware_specs}
\begin{tabular}{lccc}
\toprule
GPU & VRAM & Config. & mIoU (\%) \\ 
\midrule
RTX 5080 & 16GB & 2*Mam.+1*G-Mam. & \textbf{80.80} \\
RTX 4060Ti & 8GB & 1*G-Mamba & 77.97 \\
RTX 3060Ti & 8GB & 1*G-Mamba & 77.45 \\
\bottomrule
\end{tabular}
\end{table}

\begin{table}[t]
\centering
\caption{Comparison of semantic segmentation performance on the ADE20K validation set. All methods utilize ResNet-101 as the backbone to ensure a fair comparison of architectural efficiency. DGM-Net achieves highly competitive results against established attention-based models.}
\label{tab:ade20k_results}
\small 
\setlength{\tabcolsep}{8pt} 
\begin{tabular}{l|c|c}
\hline
Method & Venue & ADE20K (mIoU \%) \\ \hline \hline
PSPNet & CVPR'17 & 43.29 \\
PSANet & ECCV'18 & 43.77 \\
SAC    & CVPR'19 & 44.30 \\
DSSPN  & CVPR'18 & 43.68 \\
EncNet & CVPR'18 & 44.65 \\
GCU    & ECCV'18 & 44.81 \\
APCNet & CVPR'19 & 45.38 \\
CCNet  & ICCV'19 & 45.22 \\
ANN    & ICCV'19 & 45.24 \\
OCRNet & ECCV'20 & 45.28 \\ \hline
\textbf{DGM-Net (Ours)} & -- & \textbf{45.24} \\ \hline
\end{tabular}
\end{table}

\begin{figure*}[t]
  \centering
  \small
  
  \makebox[0.18\linewidth]{Ground Truth} \hfill
  \makebox[0.18\linewidth]{V-map}      \hfill
  \makebox[0.18\linewidth]{D-Map}      \hfill
  \makebox[0.18\linewidth]{Flow Field} \hfill
  \makebox[0.18\linewidth]{Inference}  \\
  \vspace{2pt}

  \includegraphics[width=0.18\linewidth]{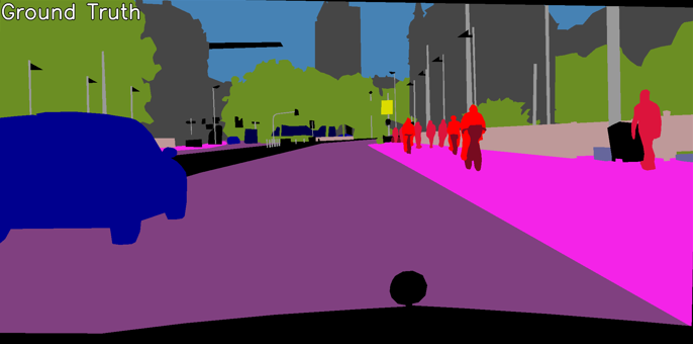} \hfill
  \includegraphics[width=0.18\linewidth]{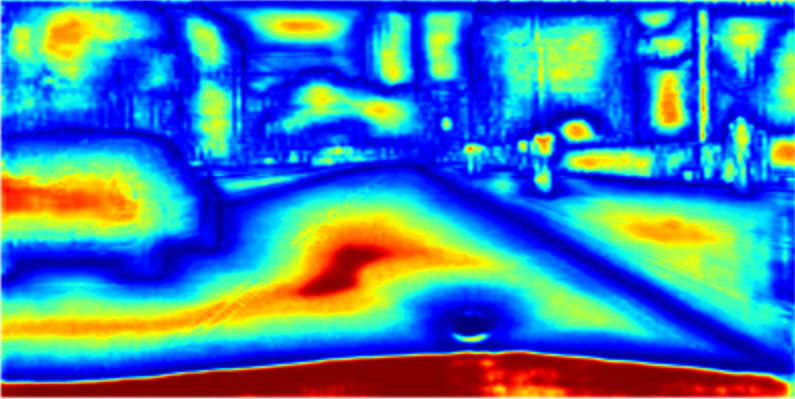} \hfill
  \includegraphics[width=0.18\linewidth]{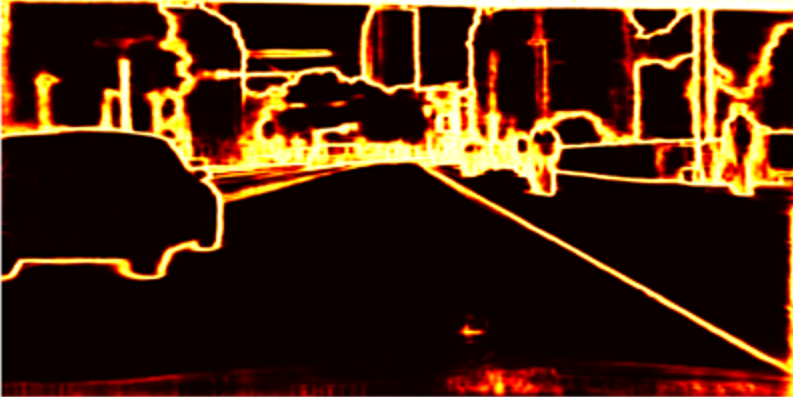} \hfill
  \includegraphics[width=0.18\linewidth]{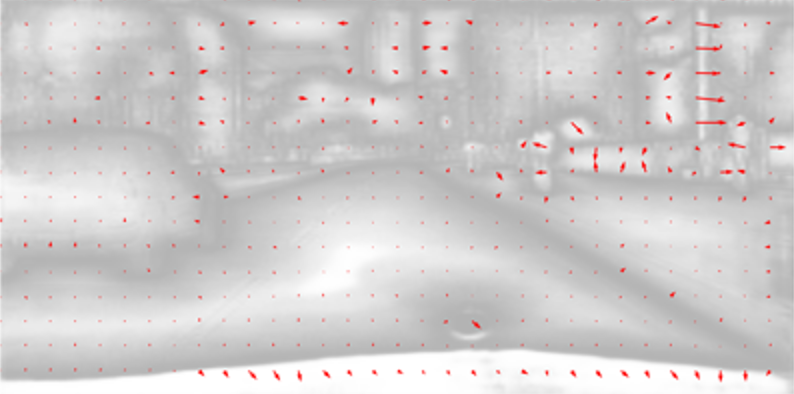} \hfill
  \includegraphics[width=0.18\linewidth]{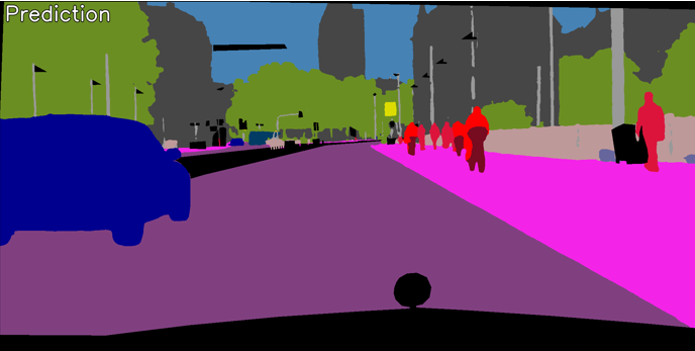} \\
  \vspace{4pt}

  \includegraphics[width=0.18\linewidth]{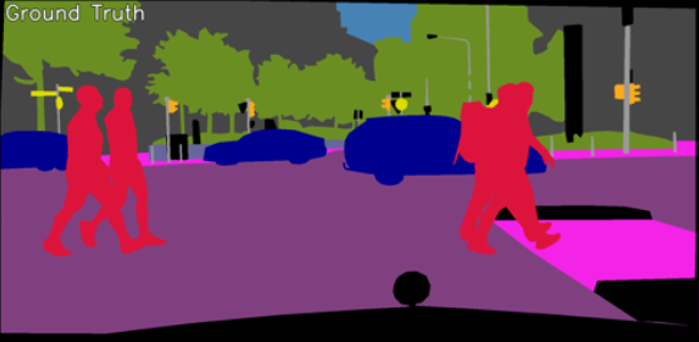} \hfill
  \includegraphics[width=0.18\linewidth]{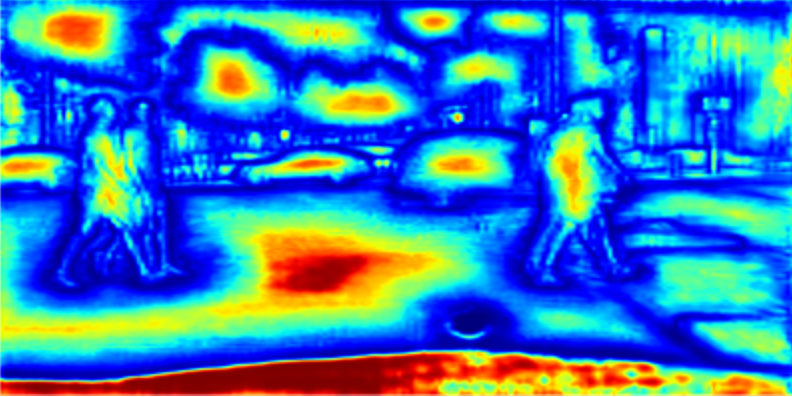} \hfill
  \includegraphics[width=0.18\linewidth]{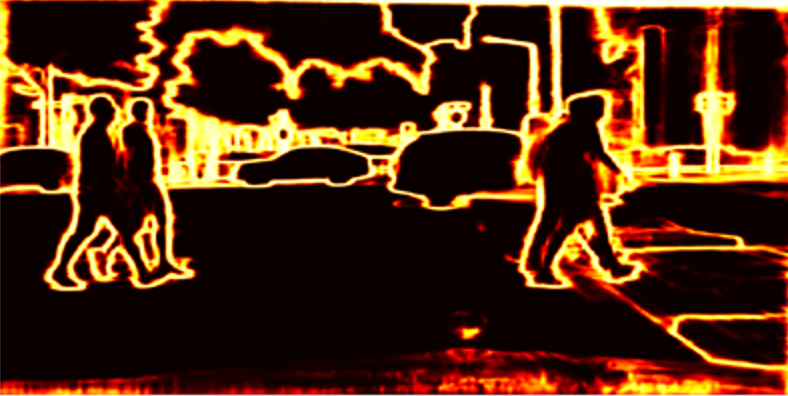} \hfill
  \includegraphics[width=0.18\linewidth]{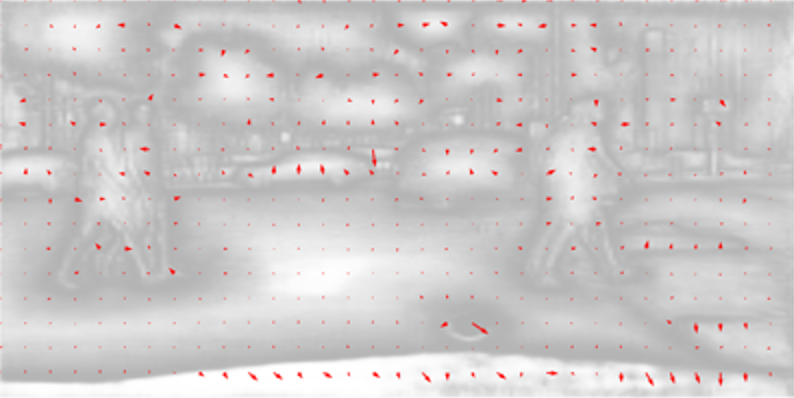} \hfill
  \includegraphics[width=0.18\linewidth]{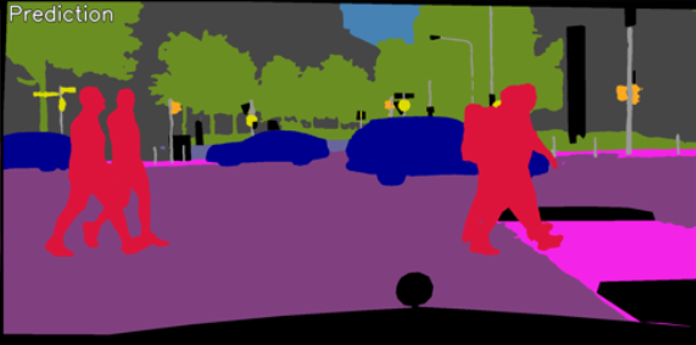} \\
  \vspace{4pt}

  \includegraphics[width=0.18\linewidth]{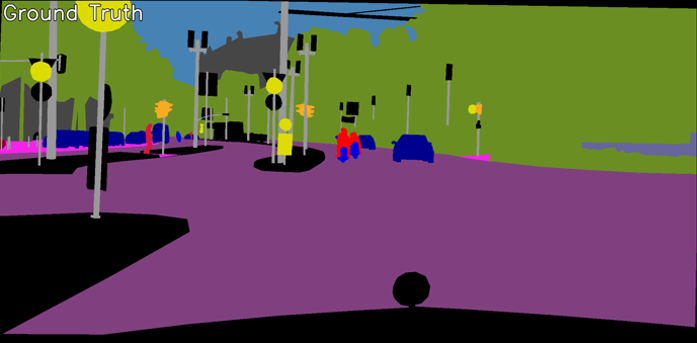} \hfill
  \includegraphics[width=0.18\linewidth]{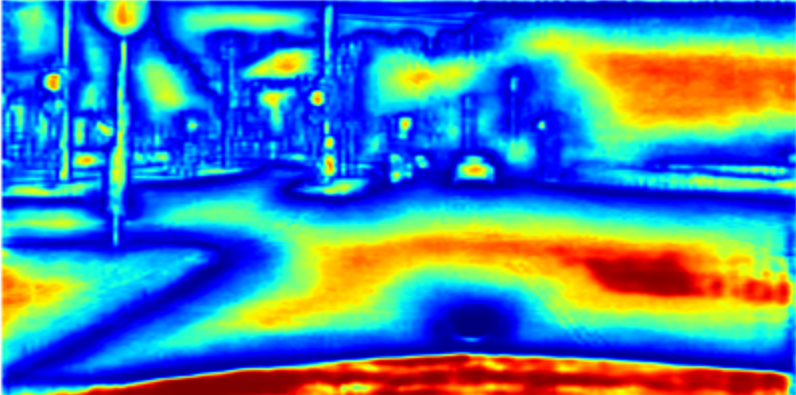} \hfill
  \includegraphics[width=0.18\linewidth]{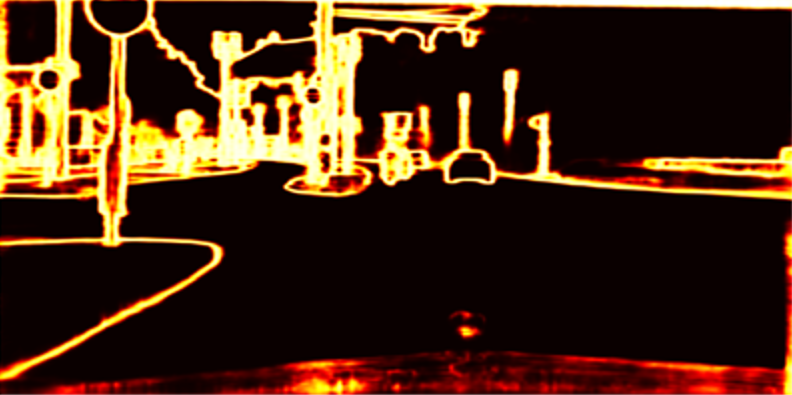} \hfill
  \includegraphics[width=0.18\linewidth]{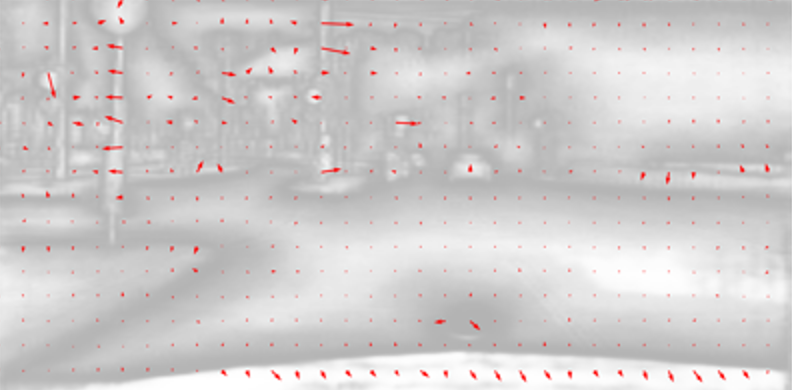} \hfill
  \includegraphics[width=0.18\linewidth]{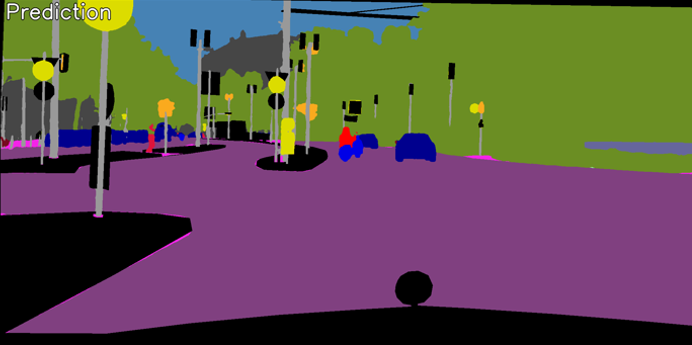} \\
  \vspace{4pt}

  \caption{Qualitative comparison on Cityscapes. Each row displays (from left to right): Ground Truth, V-map, D-Map, Flow Field, and our Inference result. Our method consistently produces sharper boundaries across different scenarios.}
  \label{fig:multi_vis_results}
\end{figure*}

\subsection{Robustness and Cross-Domain Generalization}
To evaluate the robustness and generalization capability of DGM-Net beyond urban driving scenarios, we extend our evaluation to a heterogeneous benchmark: ADE20K \cite{zhou2017scene}. This dataset presents a significant departure from the structural regularities of Cityscapes, encompassing a wide variety of indoor and outdoor environments, as well as diverse object categories that pose substantial challenges for maintaining geometric consistency.

As summarized in Table~\ref{tab:ade20k_results}, DGM-Net achieves a competitive performance of 45.24\% mIoU on the ADE20K validation set. It is important to emphasize that these results are obtained using the exact same architectural configuration and geometric prior strategies (V-map, C-map, and D-map) originally optimized for Cityscapes, without any dataset-specific structural modifications or hyperparameter tuning.

The consistent performance on ADE20K demonstrates that the geometric cues extracted by the proposed DGM-module are not limited to urban scene structures. Instead, they capture more generalizable structural representations that effectively guide the G-Mamba scanning process across diverse spatial layouts. This observation suggests that integrating geometric priors with State Space Models (SSMs) provides a robust and scalable framework for general semantic segmentation tasks.

\subsection{Visualization}

To provide qualitative evidence for the effectiveness of the proposed design, we present comprehensive visualization results on the Cityscapes validation set. First, as shown in Figure \ref{fig:multi_vis_results}, we demystify the geometric reasoning process by displaying the intermediate priors generated by our DGM-Module. By explicitly extracting the V-map, Curv-map, and Flow Field, the network is equipped with a strong morphological prompt, allowing it to produce sharper boundaries and effectively suppress semantic leakage across different complex scenarios.

\begin{figure}[!t]
  \centering
  \includegraphics[width=\linewidth,height=0.12\textheight,keepaspectratio]{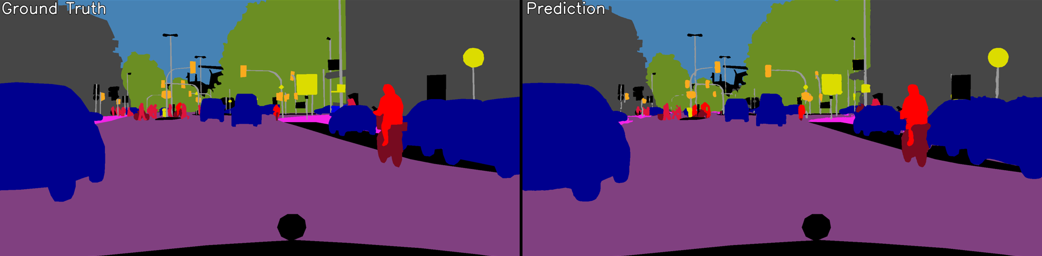}
  \includegraphics[width=\linewidth,height=0.12\textheight,keepaspectratio]{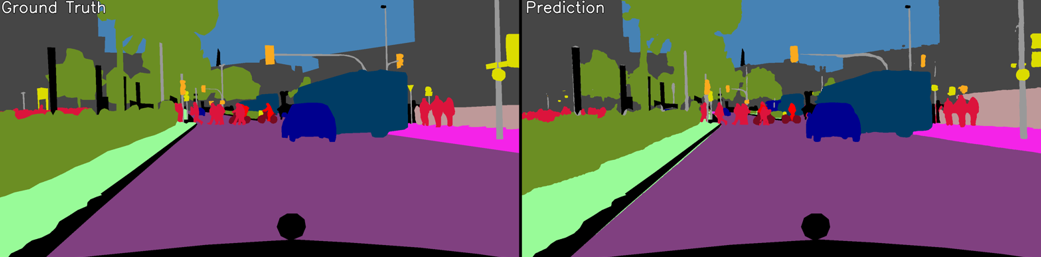}
  \includegraphics[width=\linewidth,height=0.12\textheight,keepaspectratio]{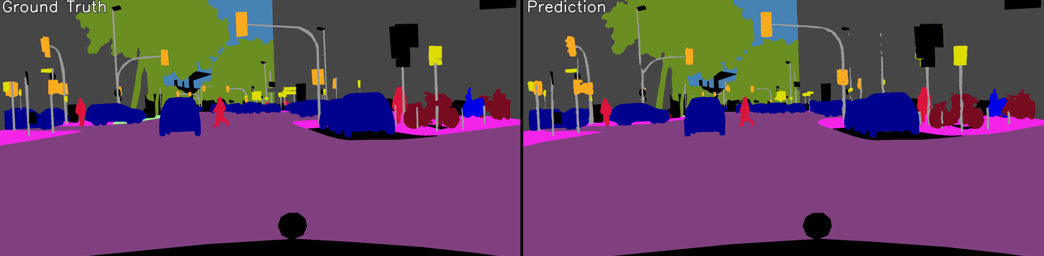}
  \includegraphics[width=\linewidth,height=0.12\textheight,keepaspectratio]{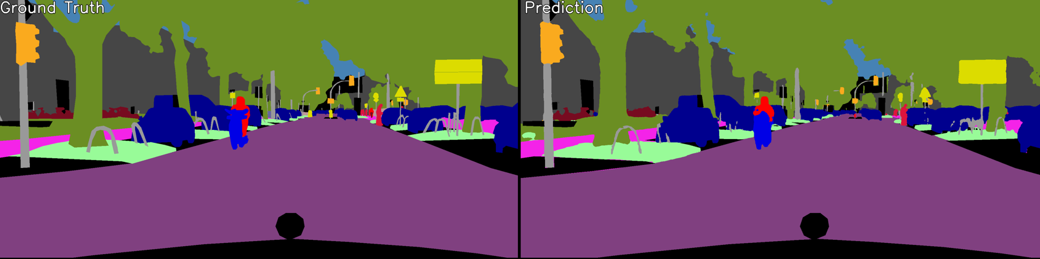}
  \includegraphics[width=\linewidth,height=0.12\textheight,keepaspectratio]{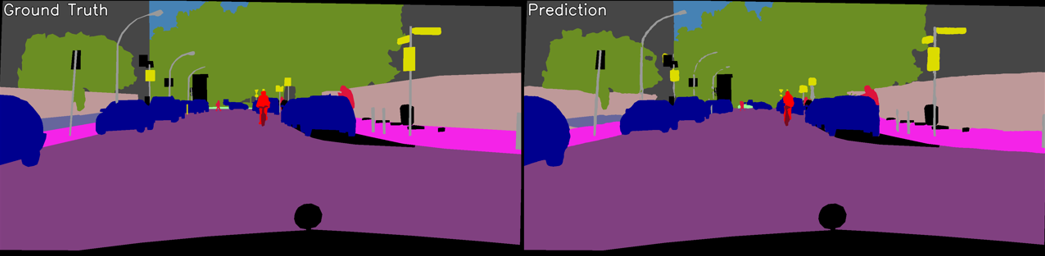}
  \vspace{1pt}
  \small \hspace{0.15\linewidth} Ground Truth \hfill Prediction \hspace{0.15\linewidth}
  \caption{Visualization results on Cityscapes val set.}
  \label{fig:vis_results}
\end{figure}

\begin{figure}[t]
  \centering
  \small
  
    \makebox[0.27\linewidth][c]{Ground Truth} \hfill
    \makebox[0.27\linewidth][c]{Inference} \hfill
    \makebox[0.27\linewidth][c]{Overlay Visualization} \\
    \vspace{2pt}

  \includegraphics[width=0.27\linewidth]{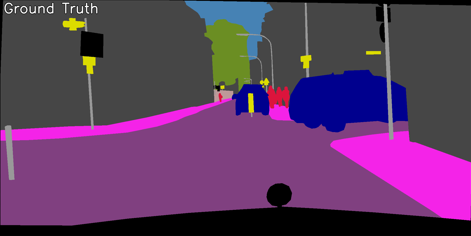} \hfill
  \includegraphics[width=0.27\linewidth]{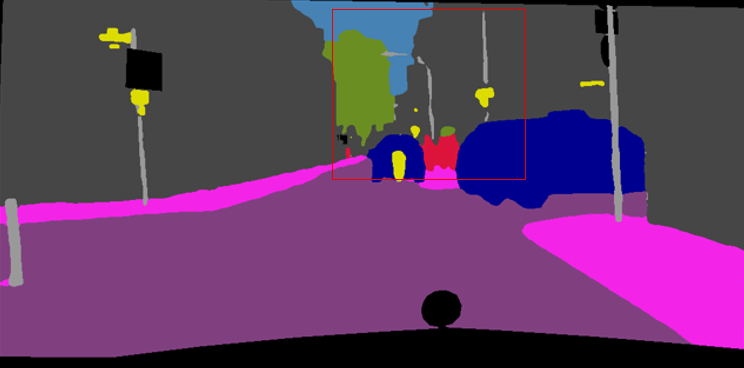} \hfill
  \includegraphics[width=0.27\linewidth]{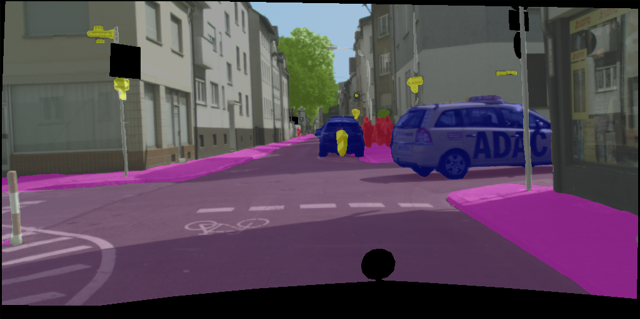} \\
  \vspace{4pt}

  \includegraphics[width=0.27\linewidth]{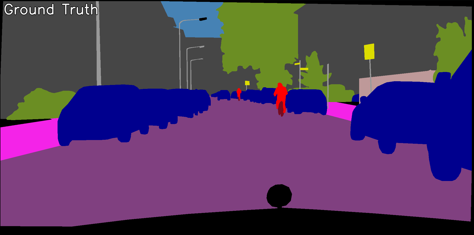} \hfill
  \includegraphics[width=0.27\linewidth]{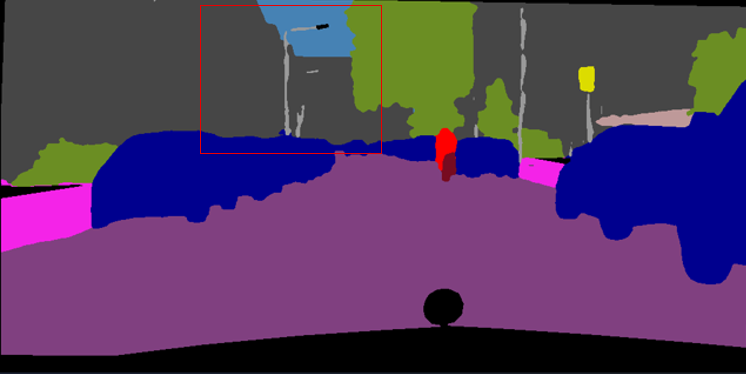} \hfill
  \includegraphics[width=0.27\linewidth]{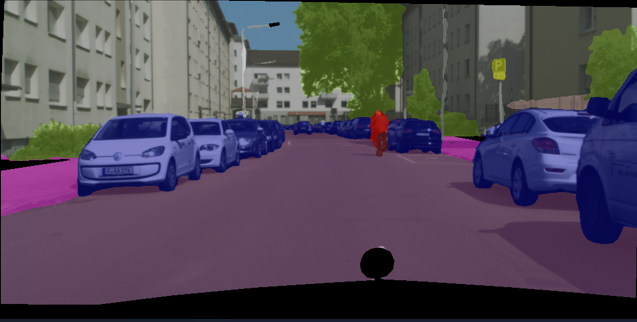} \\
  \vspace{4pt}

  \includegraphics[width=0.27\linewidth]{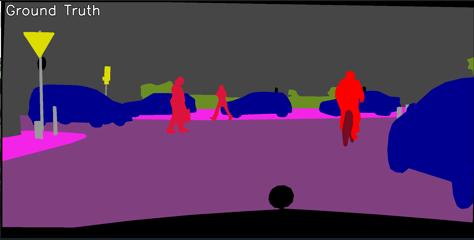} \hfill
  \includegraphics[width=0.27\linewidth]{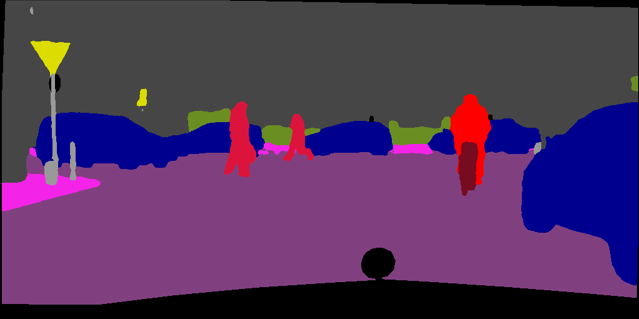} \hfill
  \includegraphics[width=0.27\linewidth]{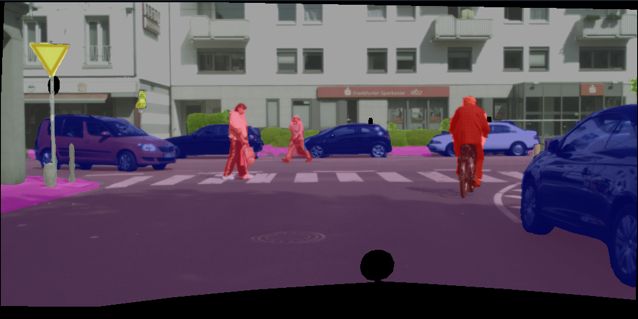} \\
  \vspace{4pt}

  \includegraphics[width=0.27\linewidth]{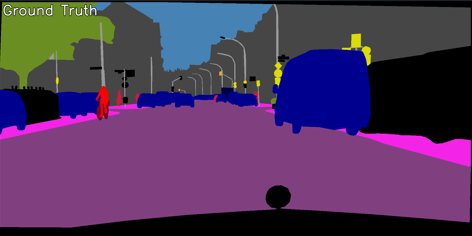} \hfill
  \includegraphics[width=0.27\linewidth]{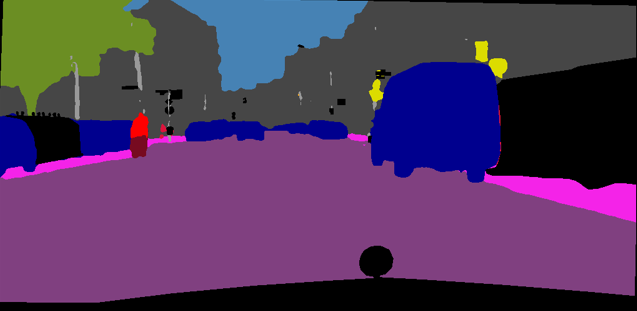} \hfill
  \includegraphics[width=0.27\linewidth]{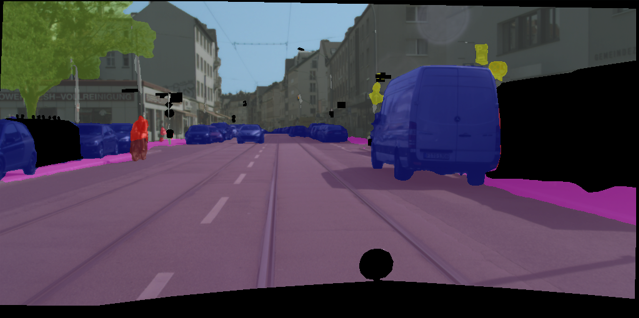} \\
  \vspace{4pt}

  \includegraphics[width=0.27\linewidth]{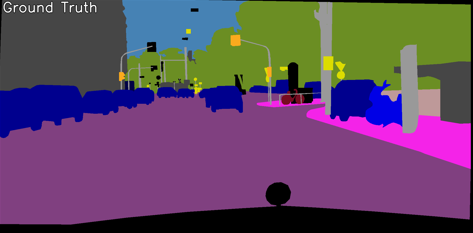} \hfill
  \includegraphics[width=0.27\linewidth]{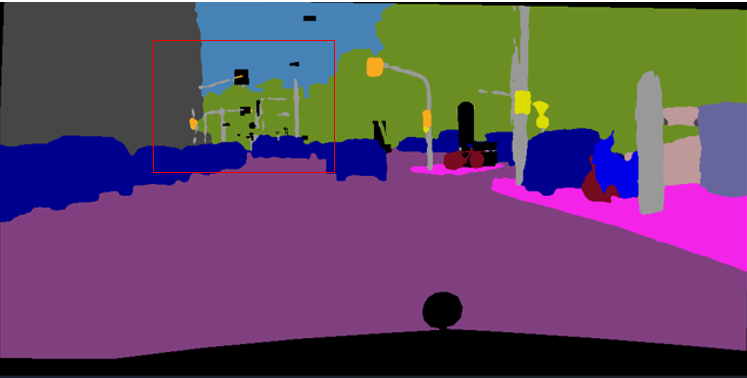} \hfill
  \includegraphics[width=0.27\linewidth]{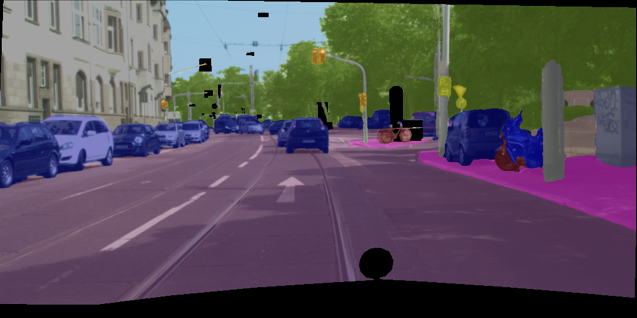} \\
  \vspace{4pt}

  \caption{Even under a constrained setting (Batch Size = 2), DGM-Net maintains structural integrity. The red circles highlight the model's ability to preserve slender poles and sharp boundaries, demonstrating that our geometric priors effectively stabilize the global context modeling of G-Mamba without requiring large-scale GPU clusters.}
  \label{fig:3060}
\end{figure}

Building upon these priors, Figure \ref{fig:vis_results} compares our final DGM-Net predictions with the ground truth. Our method consistently shows strong alignment with the ground truth annotations, particularly in preserving the structural integrity of thin and boundary-sensitive objects. To validate our claim of hardware robustness, Figure \ref{fig:3060} visualizes predictions trained under a highly constrained setting (Batch Size = 2 on a single GPU). The highlighted regions demonstrate that DGM-Net can maintain high-quality semantic context and preserve slender poles without relying on the massive batch sizes typically required by global attention mechanisms.

\begin{figure}
  \centering
  \small
  
  \makebox[0.31\linewidth]{Ground Truth} \hfill
  \makebox[0.31\linewidth]{1 $\times$ G-Mamba} \hfill
  \makebox[0.31\linewidth]{Cascade G-Mamba} \\
  \vspace{2pt}

  \includegraphics[width=0.31\linewidth]{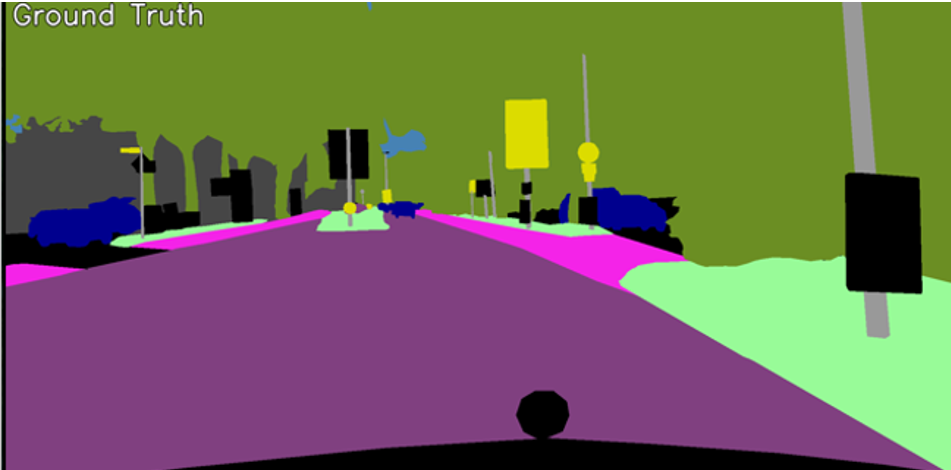} \hfill
  \includegraphics[width=0.31\linewidth]{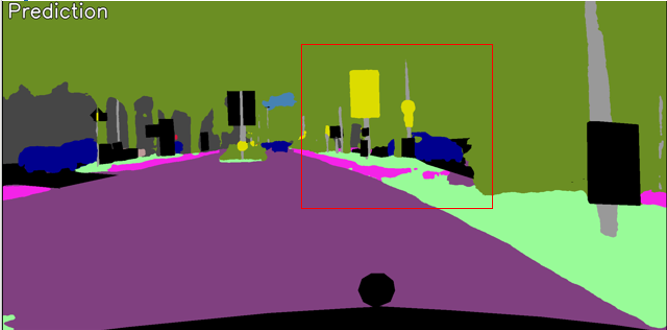} \hfill
  \includegraphics[width=0.31\linewidth]{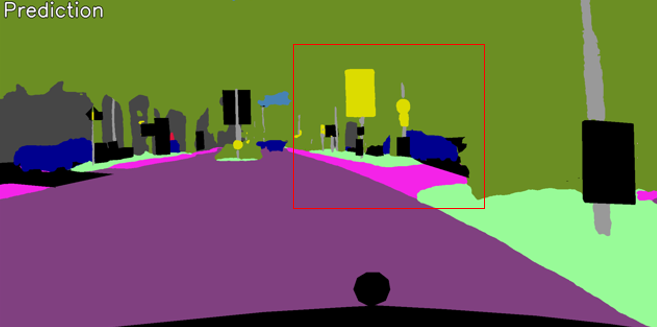} \\
  \vspace{4pt}

  \includegraphics[width=0.31\linewidth]{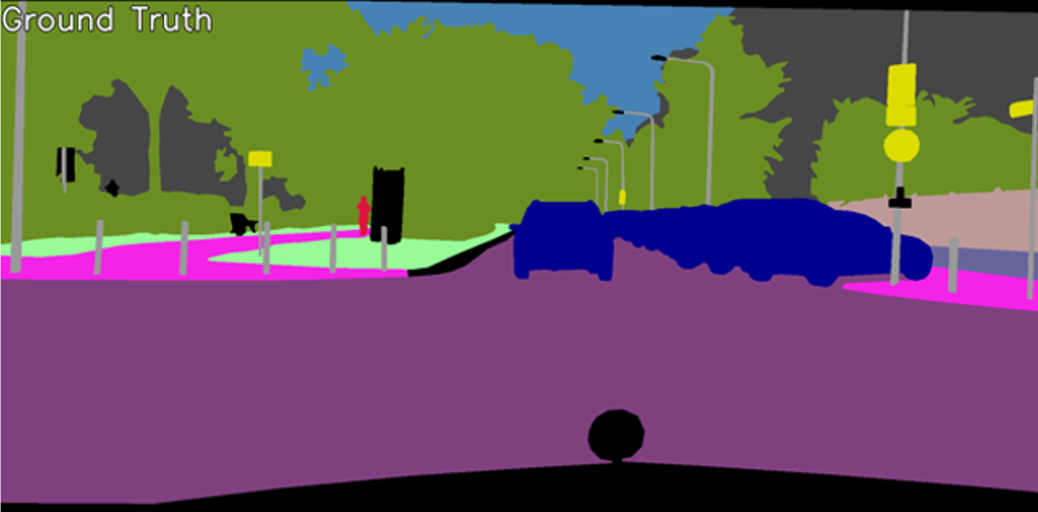} \hfill
  \includegraphics[width=0.31\linewidth]{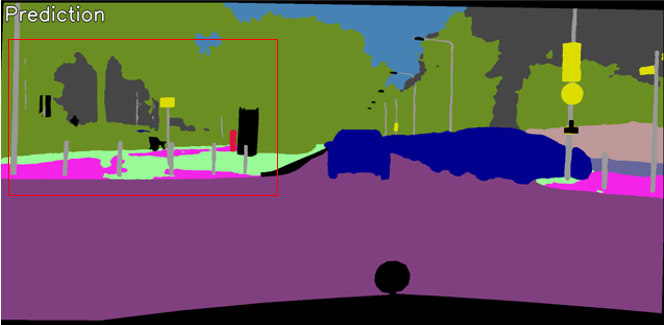} \hfill
  \includegraphics[width=0.31\linewidth]{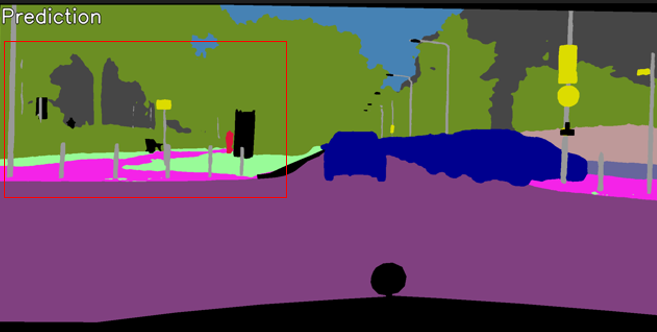} \\
  \vspace{4pt}

  \includegraphics[width=0.31\linewidth]{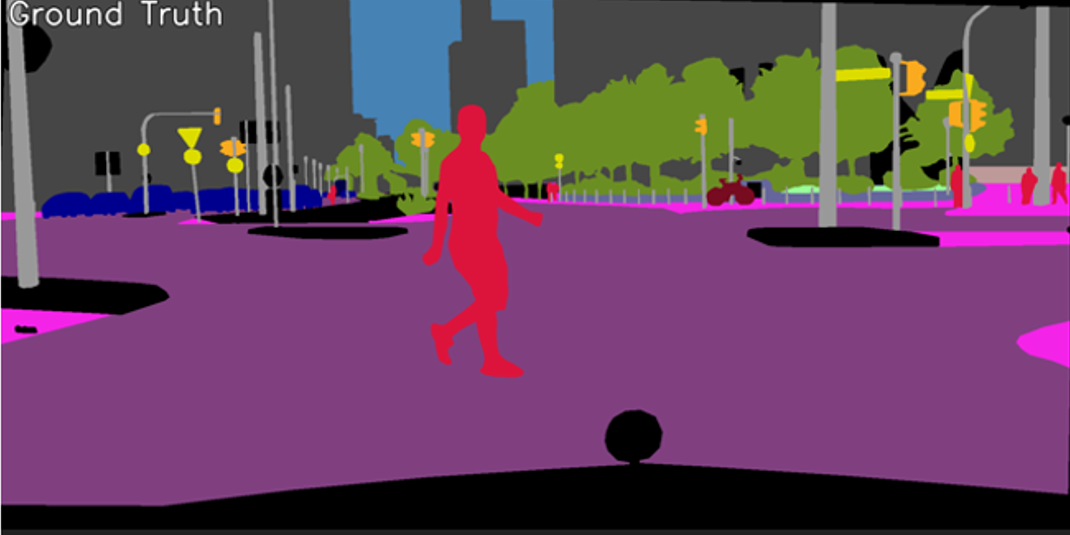} \hfill
  \includegraphics[width=0.31\linewidth]{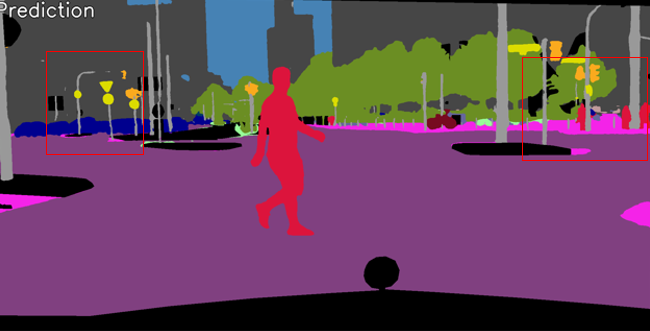} \hfill
  \includegraphics[width=0.31\linewidth]{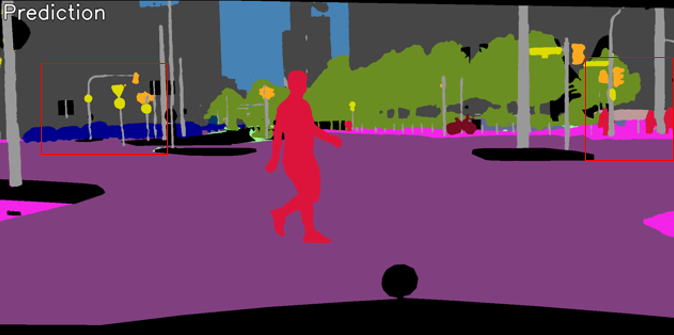} \\
  \vspace{4pt}

  \includegraphics[width=0.31\linewidth]{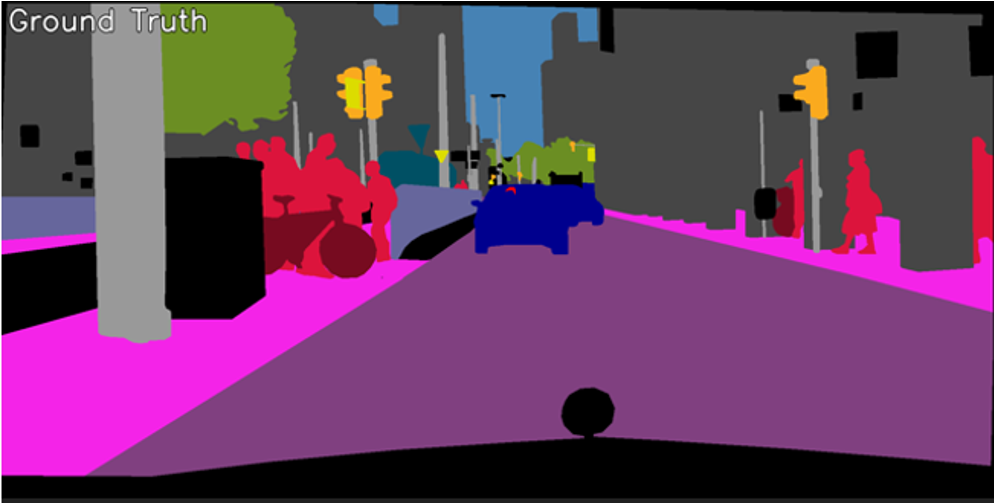} \hfill
  \includegraphics[width=0.31\linewidth]{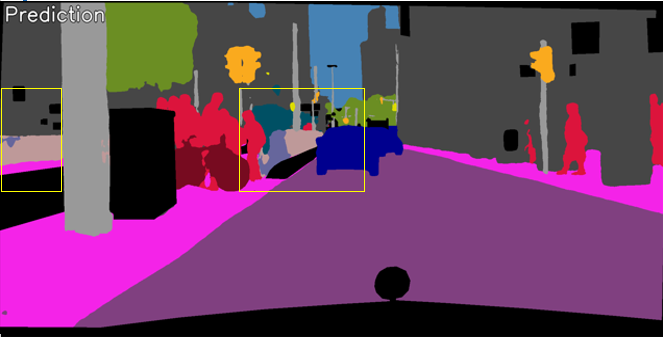} \hfill
  \includegraphics[width=0.31\linewidth]{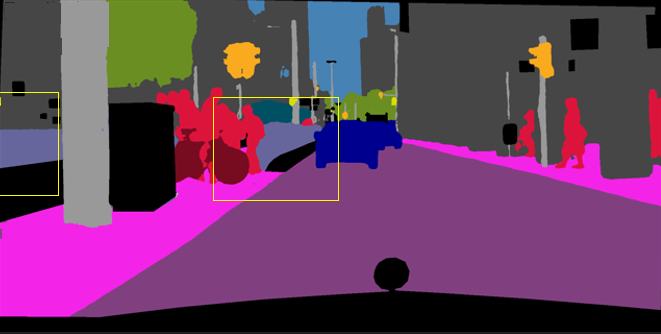} \\
  \vspace{4pt}

  \caption{It shows qualitative ablation of the cascade design. Compared to the single G-Mamba variant (middle), our full Cascade G-Mamba architecture (right) significantly reduces semantic fragmentation and produces more coherent predictions by progressively fusing global context with geometric priors.}
  \label{fig:cascade_comparison}
\end{figure}

\noindent\textbf{Visual Impact of the Cascade Architecture}\\ To further demonstrate the necessity of our progressive Cascade G-Mamba design, we qualitatively compare the predictions of a single G-Mamba variant against our full cascade architecture in Figure \ref{fig:cascade_comparison}. While a single G-Mamba layer can capture certain local geometric boundaries, it occasionally struggles with macro-semantic consistency in highly complex scenes, leading to fragmented or isolated predictions. In contrast, the Cascade G-Mamba—by establishing a robust global semantic foundation in its initial layers ($L_1, L_2$) before applying the strict geometric navigation ($L_3$)—successfully maintains semantic coherence and produces significantly cleaner, more continuous segmentation masks.

\subsection{Limitations}
\label{sec:limitations}

While DGM-Net achieves a favorable balance between accuracy and computational efficiency, several limitations remain that warrant further investigation.

First, although DGM-Net improves the representation of thin and elongated structures compared to vanilla SSM-based designs, challenges still remain in accurately modeling extremely sparse and high-frequency details. This limitation is largely attributed to the intrinsic smoothing behavior of SSM-based aggregation, which may suppress fine-grained structural variations in complex scenes.

Second, the proposed Directional Flow Field is primarily designed for “thing” classes with well-defined geometric centers. For large-scale “stuff” regions (e.g., sky, road, vegetation) that lack clear structural centroids, the effectiveness of centripetal guidance is relatively reduced, as the underlying geometric assumptions become less informative in such scenarios.

Finally, as our primary focus is on efficient single-GPU training, DGM-Net is evaluated without large-scale pretraining. While this setting highlights its strong learning capability under constrained computational budgets, the scalability and performance potential under large-scale data regimes remain an important direction for future exploration.
\section{Conclusion}
\label{sec:conclusion}
In this paper, we propose DGM-Net, an efficient semantic segmentation framework that integrates State Space Models with geometry-guided priors. Instead of increasing model capacity or relying on large-scale pretraining, our approach focuses on improving structural awareness and feature propagation under constrained computational resources. Experimental results on Cityscapes and ADE20K show that DGM-Net achieves competitive performance while maintaining efficient training and inference. In particular, the model demonstrates stable convergence behavior and maintains robustness under limited hardware settings. These findings suggest that incorporating geometric guidance into SSM-based architectures can serve as a practical direction for balancing accuracy and efficiency in semantic segmentation.\\\\
\noindent\textbf{Acknowledgments}\\
This research work is partially supported by National Science and Technology Council, Taiwan, under grant number: 114-2221-E-032-011-

{
    \small
    \bibliographystyle{plain}
    \bibliography{main}
}


\end{document}